\documentclass[12pt]{spieman}  
\usepackage{amsmath,amsfonts,amssymb}
\usepackage{graphicx}
\usepackage{caption}
\usepackage{subcaption}
\usepackage{setspace}
\usepackage{tocloft}
\usepackage{multirow}
\usepackage{booktabs}
\usepackage[table,xcdraw]{xcolor}
\usepackage{soul} 
\usepackage{hyperref}
\usepackage{listings}
\usepackage{sidecap}
\usepackage{lineno}

\title{Automatic Landmarks Correspondence Detection in Medical Images with an Application to Deformable Image Registration}

\author[a,*]{Monika Grewal}
\author[b]{Jan Wiersma}
\author[b]{Henrike Westerveld}
\author[a,c]{Peter A. N. Bosman}
\author[d]{Tanja Alderliesten}
\affil[a]{Evolutionary Intelligence Research Group, Centrum Wiskunde \& Informatica, 1098 XG, Amsterdam, The Netherlands}
\affil[b]{Department of Radiation Oncology, Amsterdam University Medical Centers, location AMC, University of Amsterdam, 1105 Amsterdam, The Netherlands}
\affil[c]{Faculty of Electrical Engineering, Mathematics and Computer Science, Delft University of Technology, 2600 AA Delft, The Netherlands}
\affil[d]{Department of Radiation Oncology, Leiden University Medical Center, 2300 RC Leiden, The Netherlands}

\cftpagenumbersoff{figure}
\cftpagenumbersoff{table} 
\begin{document} 
\maketitle

\begin{abstract}
\\\textbf{Purpose}: Deformable Image Registration (DIR) can benefit from additional guidance using corresponding landmarks in the images. However, the benefits thereof are largely understudied, especially due to the lack of automatic landmark detection methods for three-dimensional (3D) medical images.
\\\textbf{Approach}: We present a Deep Convolutional Neural Network (DCNN), called DCNN-Match, that learns to predict landmark correspondences in 3D images in a self-supervised manner. We trained DCNN-Match on pairs of Computed Tomography (CT) scans containing simulated deformations. We explored five variants of DCNN-Match that use different loss functions and assessed their effect on the spatial density of predicted landmarks and the associated matching errors. We also tested DCNN-Match variants in combination with the open-source registration software Elastix to assess the impact of predicted landmarks in providing additional guidance to DIR.
\\\textbf{Results}: We tested our approach on lower-abdominal CT scans from cervical cancer patients: 121 pairs containing simulated deformations and 11 pairs demonstrating clinical deformations. The results showed significant improvement in DIR performance when landmark correspondences predicted by DCNN-Match were used in the case of simulated (p = $0e^0$) as well as clinical deformations (p = $0.030$). We also observed that the spatial density of the automatic landmarks with respect to the underlying deformation affect the extent of improvement in DIR. Finally, DCNN-Match was found to generalize to Magnetic Resonance Imaging (MRI) scans without requiring retraining, indicating easy applicability to other datasets.
\\\textbf{Conclusions}: DCNN-Match learns to predict landmark correspondences in 3D medical images in a self-supervised manner, which can improve DIR performance.
\end{abstract}

\keywords{Deformable Image Registration, Computed Tomography, landmarks detection, Deep Learning}

{\noindent \footnotesize\textbf{*}Monika Grewal,  \linkable{monika.grewal@cwi.nl} }

\begin{spacing}{2}   

\section{Introduction}
\label{sect:intro}  

Deformable Image Registration (DIR) is a task of aligning a source (or moving) image to a target (or fixed) image by optimizing a Displacement Vector Field (DVF). The aligned source image can then be computed by resampling the source image at the spatial locations specified by the mapping. DIR has tremendous application possibilities in the radiation treatment workflow required for cancer treatment e.g., automatic contour propagation \cite{chao2008auto, ghose2015review}, dose accumulation \cite{thor2014evaluation, rigaud2019deformable, chetty2019deformable}. However, DIR in regions such as the pelvis is challenging due to large local deformations and appearance differences caused by physical processes such as bladder filling, and the presence of gas pockets and contrast agents \cite{ghose2015review}. In such DIR scenarios, the existing non-linear intensity-based registration approaches \cite{KleinStaring:2010, vercauteren2009diffeomorphic, weistrand2015anaconda} often get stuck in a local minimum\cite{rigaud2019deformable}. Many previous studies \cite{alderliesten2015getting, Werner2013, polzin2013combining, ForstnerOperator, hervella2018multimodal, HAN2015277} have shown that landmark correspondences between the images to be registered can provide additional guidance to the intensity-based DIR methods and help overcome local minima. However, to the best of our knowledge, such an approach has not been tested on pelvic scans.

Manual annotation of landmarks for DIR in the clinic is not practically tractable due to two main reasons. First, a high number of landmarks is desired, and it is difficult to unambiguously define such a high number of landmarks manually. Second, manual annotations require lots of time from clinicians, which is hardly available. Therefore, an automatic method for  finding landmark correspondences is required. Although many endeavours have been made in the direction of automatic landmarks correspondence detection in medical images \cite{yang2017method, HAN2015277, XRayLandmarks}, there remain significant gaps to fill. The existing methods usually employ large pipelines consisting of multiple components, each component using multiple hyperparameters derived from image features specific to the underlying dataset. Consequently, the entire pipeline is sensitive to small variations in local image intensities and choices of hyperparameters, making application to a new dataset difficult. Moreover, in datasets such as pelvic scans with ill-defined boundaries between soft tissues, intensity gradient based landmark detection may not work at all. 

Convolutional Neural Networks (CNNs) are known to learn deep features from images, which are robust to small variations in local image intensities. In recent years, deep CNNs have not only shown remarkable performance in difficult computer vision tasks in medical imaging \cite{JAMARetinopathy, esteva2017dermatologist}, but also good generalization to unseen data. Moreover, with the advances in the available computational resources, CNN-based solutions turn out to be faster than their traditional counterparts. Therefore, there is a strong motivation to replace the entire pipeline for automatically finding landmark correspondences by a deep CNN. Recently some deep CNN methods have been developed for automatic landmark detection in medical images \cite{tuysuzoglu2018deep, artificialagent, GrewalLandmark}, but these are limited to either 2D datasets or supervised learning of a few manually annotated landmarks. Other relevant works include methods for landmark propagation from a template image by learning pixel-wise anatomical embeddings \cite{yan2020self} or through deformable image registration \cite{devine2020registration}. While such methods allow for single shot landmark detection in a new image, the requirement of manual annotation of landmarks on the template image still exists. Another study uses unsupervised image registration as a proxy task to discover landmarks shape descriptors \cite{bhalodia2021leveraging}, but this method is limited to discovering a small number of landmarks ($\sim$ 100 landmarks per image pair).

In this study, we present a deep CNN (referred to as ``DCNN-Match") for automatic landmarks correspondence detection (i.e., simultaneous landmark detection as well as matching) in 3D images. The presented method is an extension of our method for 2D images \cite{GrewalLandmark}. Briefly, the neural network is trained on pairs of 3D lower abdominal Computed Tomography (CT) scans such that the network learns to predict landmarks at salient locations in both the images along with the correspondence score of each landmark pair. One key feature of the presented method is that unlike supervised methods, the neural network in the presented method is trained in a self-supervised manner without using any manual annotations. This is important because manual annotations on medical images are not always readily available, mainly because it is time-consuming to create them.

It is essential to investigate the added value of automatic landmarks correspondence detection towards the improvement of the DIR solutions to estimate the potential deployability of landmarks-guided DIR approaches in the clinic. Existing studies have investigated the added value of automatic landmark correspondences towards DIR independently of the underlying automatic landmark detection method \cite{Werner2013, polzin2013combining, HAN2015277}. Since change in the automatic landmarks correspondence detection method changes the aspects of the automatic landmarks e.g., spatial distribution and matching accuracy, the effect of the automatic landmarks on the DIR performance is likely to be affected as well. Therefore, we believe that developing a method for automatic landmarks correspondence detection and at the same time integrating it with a DIR pipeline can provide numerous insights. To this end, we have integrated our method for automatic landmark detection and matching with an existing DIR software so that the added value of using landmark correspondences in solving DIR problems can be assessed. Further, we investigate five different variants of the developed method by use of different loss functions during training that each predict landmark correspondences with different spatial distributions and matching errors, to assess the effect of different types of automatic landmark correspondences towards the improvement of DIR.
The present work has the following contributions:
\begin{itemize}
    \item We extended our previously published end-to-end self-supervised deep learning method for automatically finding landmark correspondences in medical images from 2D to 3D. The key highlights of the method are:
    \begin{itemize}
        \item the method does not set any prior on the definition of landmarks
        \item the method does not require manual annotations for training
    \end{itemize}
    \item We integrated our automatic landmark correspondence detection method in 3D (DCNN-Match) with an open-source registration software Elastix \cite{Elastix2:2014, KleinStaring:2010} to develop a DIR pipeline that utilizes additional guidance information from automatic landmark correspondences. We used this DIR pipeline to investigate the added value of automatic landmark correspondences in providing additional guidance to the DIR method and finding better DIR solutions.
    \item We varied the landmarks correspondence detection method and investigated how it affected the added value to the DIR method. We explored five different variants of the proposed automatic landmarks correspondence method.
    \item We experimentally investigated the generalization capability of our proposed automatic landmarks correspondence detection method to a Magnetic Resonance Imaging (MRI) dataset.
\end{itemize}

\section{Materials and Methods}
In the following sections, we describe DCNN-Match (section \ref{sec:DCNN-Match}), and the DIR pipeline which uses the information from automatic landmark correspondences predicted by DCNN-Match to guide the registration (section \ref{sec:DIR pipeline}). Sections \ref{sec:implementation} and \ref{sec:hyperparameters} provide details of implementation and hyperparameters for reproducibility. Sections \ref{sec:data}, \ref{section:experiments}, \ref{sec:evaluation}, and \ref{sec:statistics} describe the datasets, experiments, evaluation metrics, and statistical testing used in the experiments, respectively.

\subsection{DCNN-Match}
\label{sec:DCNN-Match}

\begin{figure*}[h!]
    \centering
    \includegraphics[width=0.95\textwidth]{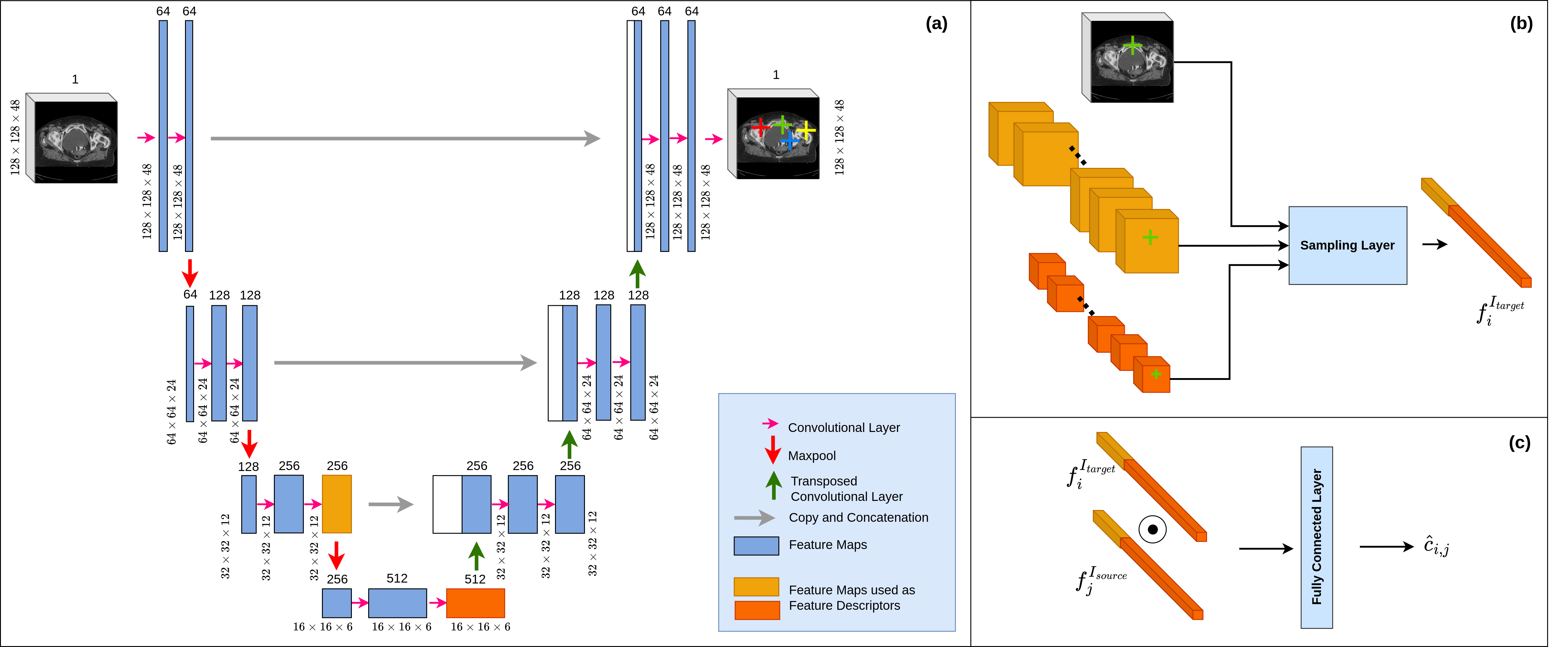}
    \caption{\textbf{Illustration of the components of DCNN-Match.} (a) Illustration of different layers in the shared \textbf{CNN branch} used for landmark detection and feature description. (b) The \textbf{sampling layer} samples the feature maps of the last two downsampling levels in the CNN branch at the locations described by the landmark probability map. (c) The \textbf{descriptor matching module} realized by a fully connected layer predicts the matching probability of a feature descriptor pair.}
    \label{fig:DCNN-Match}
\end{figure*}

We extended our approach \cite{GrewalLandmark} for finding landmark correspondences in 2D CT scan slices to work on 3D CT scans. The different components of the 3D approach are illustrated in Figure \ref{fig:DCNN-Match}. 
Briefly, the approach proposed in \cite{GrewalLandmark} consists of a Siamese network with three modules: a) two CNN branches with shared weights, b) a sampling layer, c) a descriptor matching module. The CNN branches comprise an image-to-image translation network that maps an input image to a feature map. The architecture of the network is derived from the famous UNet architecture \cite{UnetMiccai} proposed for image segmentation. For a given pair of target image ($I_{target}$) and source image ($I_{source}$), the CNN branches predict a landmark probability map describing the probability $\hat{p}^{I_x}_i$ ($x \in \{target, source\}$) of each spatial location $i$ being a landmark. The sampling layer is a parameter-free module that samples $K$ (hyperparameter) landmark locations with top landmark probabilities during training. During inference, the sampling layer samples all landmark locations with landmark probabilities above a threshold. We used the value 0.5, same as in \cite{GrewalLandmark}. 

Additionally, the sampling layer samples a feature vector from the feature maps of the last two downsampling levels in the CNN branch at the coordinates of each $i^{th}$ landmark location and constructs the feature descriptor $f^{I_x}_i$ by concatenating the sampled feature vectors. This allows for efficient use of the network weights by simultaneous learning the landmark detection as well as feature description of each landmark without unnecessarily increasing the network size. Moreover, the concatenation of features from different downsampling levels emulates the behavior of multi-scale feature description, which otherwise, is achieved by calculating features from a Gaussian pyramid representation of the image. Following the calculation of feature descriptors for each landmark location, the sampling layer creates feature descriptor pairs $(f^{I_{target}}_i, f^{I_{source}}_j) \,  \forall \,  i=1, 2, ..., K \; \text{in} \; I_{target} \;  \text{and} \; \forall \, j=1, 2,..., K \; \text{in} \; I_{source}$ to feed to the descriptor matching module. The descriptor matching module predicts the landmark matching probabilities corresponding to each feature descriptor pair.

\subsubsection{Self-supervised Training}

The network is trained in a self-supervised manner on pairs of target ($I_{target}$) and source ($I_{source}$) lower abdominal CT scans containing simulated deformations. The details on the generation of target and source image pairs are provided in section \ref{sec:implementation}.

Following the sampling of landmark locations $i = 1, 2, ..., K$ in $I_{target}$ and $j = 1, 2, ..., K$ in $I_{source}$ along with their corresponding feature descriptors $f^{I_{target}}_i$ and $f^{I_{source}}_j$, feature descriptor pairs $(f^{I_{target}}_i, f^{I_{source}}_j)$ are constructed in the sampling layer. The feature descriptor pairs are considered corresponding to all $i$ and $j$, allowing for feature descriptor matching between far-away locations in the images without requiring encoding of the underlying deformation field explicitly. Since the simulated deformations used to create source and target image pairs during training can not represent the complex large deformations in a clinical setup exactly, learning the feature descriptor matching not explicitly dependent on the underlying deformation field is likely to help the neural network generalize better to clinical scenario.

The ground truth $c_{i,j}$ of the correspondence of each feature descriptor pair is calculated on-the-fly based on the known simulated deformation. Each sampled landmark location in the target image is projected onto the source image based on the known simulated deformation and the nearest predicted landmark (within a distance of 2 voxels = 4 mm) in the source image is considered its match. We used a threshold of 4 mm (instead of image resolution = 2 mm) in order to find a reasonable number of landmark matches from random predictions in the beginning of the training to ensure sufficient supervision. the value of $c_{i,j} = 1$ for matching and $c_{i,j} = 0$ for non-matching feature descriptor pairs. Subsequently, the ground truth $p^{I_x}_i$ for the landmark probability of landmark location $i$ in image $I_x, x \in \{target, source\}$ is determined as follows:
\begin{equation}\label{eq:groundtruth}
  p^{I_x}_i=
  \begin{cases}
    1 & \text{if $\exists! \, j \in \{0, 1, 2, ..., K\}$ in image $I_y, y \in \{target, source\}, \, y!=x \land c_{i,j} = 1 $}\\
    0 & \text{otherwise}
  \end{cases}
\end{equation}
The ground truths $c_{i,j}$ are used directly as ground truths for the matching probability of the feature descriptor pairs $(f^{I_{target}}_i, f^{I_{source}}_j)$. 
In other words, the ground truth is generated such that the landmark probability as well as the descriptor matching probability is high for the matching locations between the two images and low otherwise. The network is trained by minimizing a multi-task loss defined as follows:

\begin{multline}\label{eq:loss}
    Loss = {LandmarkProbabilityLoss}_{I_{target}} + {LandmarkProbabilityLoss}_{I_{source}} \\ + {DescriptorMatchingLoss}
\end{multline}
The $LandmarkProbabilityLoss_{I_x}$ for the probabilities of landmarks in image $I_x, x \in \{target, source\}$ is defined as:
\begin{equation}\label{eq:LandmarkProbabilityLoss}
        {LandmarkProbabilityLoss}_{I_x} = \frac{1}{K} \sum_{i=1}^{K}\left((1 - \hat{p}^{I_x}_i) + CrossEntropyLoss(\hat{p}^{I_x}_i, p^{I_x}_i)\right)
\end{equation}
where $CrossEntropyLoss$ is the cross entropy loss between predicted landmark probability $\hat{p}^{I_x}_i$ and ground truth $p^{I_x}_i$ of the $i^{th}$ sampled location. $K$ is the total number of sampled landmark locations in image $I_x$. Further details of the $LandmarkProbabilityLoss$ are omitted for brevity and can be found in \cite{GrewalLandmark}.\\
The $DescriptorMatchingLoss$ allows the network to learn feature descriptor matching automatically and is defined as follows:
\begin{equation}\label{eq:DescriptorMatchingLoss}
DescriptorMatchingLoss = DescriptorHingeLoss + DescriptorCELoss
\end{equation}
$DescriptorHingeLoss$ is defined as follows:

\begin{multline}\label{eq:hingeloss}
DescriptorHingeLoss
= \sum_{i=1, j=1}^{K, K} \left( \frac{c_{i, j} max(0, ||f^{I_{target}}_i - f^{I_{source}}_j||^2 - m_{pos})}{K_{pos}} \right. \\
         + \left. \frac{(1 - c_{i, j}) max(0, m_{neg} - ||f^{I_{target}}_i - f^{I_{source}}_j||^2)}{K_{neg}}\right)
\end{multline}
where, $f^{I_{target}}_i$ and $f^{I_{source}}_j$ are the feature descriptors corresponding to the $i^{th}$ and $j^{th}$ landmark locations in the input images $I_{target}$ and $I_{source}$, respectively; $c_{i, j}$ is the ground truth matching probability for the feature descriptor pair $(f^{I_{target}}_i, f^{I_{source}}_j)$; $m_{pos}$ and $m_{neg}$ are the margins for the L2-norm of matching (positive class) and non-matching (negative class) feature descriptor pairs. The Hinge losses corresponding to positive and negative classes are normalized by $K_{pos}$ (number of positive feature descriptor pairs) and $K_{neg}$ (number of negative feature descriptor pairs), respectively to account for the class imbalance between positive and negative feature descriptor matches.
$DescriptorCELoss$ is defined as follows:
\begin{equation}\label{eq:CE}
DescriptorCELoss = \sum_{i=1, j=1}^{K, K}\left( \frac{WeightedCrossEntropy(\hat{c}_{i, j}, c_{i, j})}{(K_{pos} + K_{neg})}\right)    
\end{equation}
where $\hat{c}_{i, j}$ is the predicted matching probability; $WeightedCrossEntropy$ represents the binary cross entropy loss where the loss corresponding to the positive class is weighted by the frequency of negative examples and vice versa.

In the beginning of the training, the predicted landmark probability maps by the CNN branches are random and by chance only a few landmark locations have correct correspondence (i.e., $c_{i,j} = 1$) between images. The loss defined in \eqref{eq:loss} encourages high landmark probability at these locations as well as high feature descriptor matching probability for the feature descriptor pairs of these locations and low landmark probability and feature descriptor matching probability otherwise. Additionally, the term $(1 - \hat{p}^{I_x}_i)$ in \eqref{eq:LandmarkProbabilityLoss} encourages high landmark probability at all locations i.e., encourages more landmark locations to have correct correspondence in the other image. Consequently as the training progresses, the network learns to identify salient locations in the images that have correct correspondence in the other image as well and predicts high landmark probabilities at the these locations.

\subsubsection{End-to-end}

The conventional approach to establish landmark correspondences between an image pair utilizes the following steps:
\begin{itemize}
    \item Landmark detection, in which landmarks are detected in both the images independently.
    \item Feature description, wherein a vector (often called ``descriptor") is calculated to describe the image properties surrounding the landmark location. An example of a feature descriptor is Scale Invariant Feature Transform (SIFT \cite{lowe2004distinctive}), which calculates the histograms of orientations from the image patches of different scales around the landmark.
    \item Landmark matching, wherein landmark descriptors in both the images are matched using a matching algorithm. A straightforward matching algorithm is brute force matching, which aims at finding the best match among all the landmark locations in the source image for each landmark location in the target image.
\end{itemize}

Our approach replaces each of the abovementioned components with a neural network module, and connects the neural network modules such that the gradients flow from the end to the inputs. The modules of landmark detection and description are represented by the CNN branches of the Siamese network. The task of landmark matching is performed by the  descriptor matching module. It is important to mention that the key feature of DCNN-Match lies in the assembling of different modules to provide a simple end-to-end deep learning solution for simultaneous landmark detection, description, and matching automatically. Therefore, the proposed approach can be easily modified, e.g., it may be improved by the use of a different neural network in any of the modules.

\subsubsection{Extension to 3D images}

We have extended our original approach proposed in \cite{GrewalLandmark} to work on 3D images by performing three modifications. The first obvious modification was to use 3D convolutional kernels (kernel size =  3 $\times$ 3 $\times$ 3) instead of 2D convolutional kernels in the CNN branches. The sampling layer and the feature descriptor matching module were also adapted for 5D tensors arising from training on 3D images. The generation of a \textit{valid mask} during training as described in \cite{GrewalLandmark} section 2.4 was also adapted for 3D images. The \textit{valid mask} makes the network learn a content-based prior to predict landmarks only in the regions that include patient anatomy and not in the background or the CT couch.

Second, since we had a considerably large training dataset (details in section \ref{sec:data}) as opposed to \cite{GrewalLandmark}, we kept the same number of kernels in each layer as the original UNet architecture \cite{UnetMiccai}.
Third, we trained the network on 3D patches of the entire CT due to GPU memory constraints. During inference, we evaluated the network on the patches belonging to the same spatial locations in the target and source images. The patches were cut with 50\% overlap and the final output combined the predicted landmark pairs in all patches. All the corresponding landmarks predicted in all the overlapping patches were considered landmarks. Using a small patch size restricts the network from learning landmark matches in locations that are far apart in the two images. Therefore, the patch size has to be decided while keeping in mind the spatial extent of deformations we want the network to learn. This is further described in the hyperparameters section \ref{sec:hyperparameters}.

\subsection{DIR with Additional Guidance from Automatic Landmark Correspondences}
\label{sec:DIR pipeline}
\begin{figure*}[h!]
    \centering
    \includegraphics[width=\textwidth]{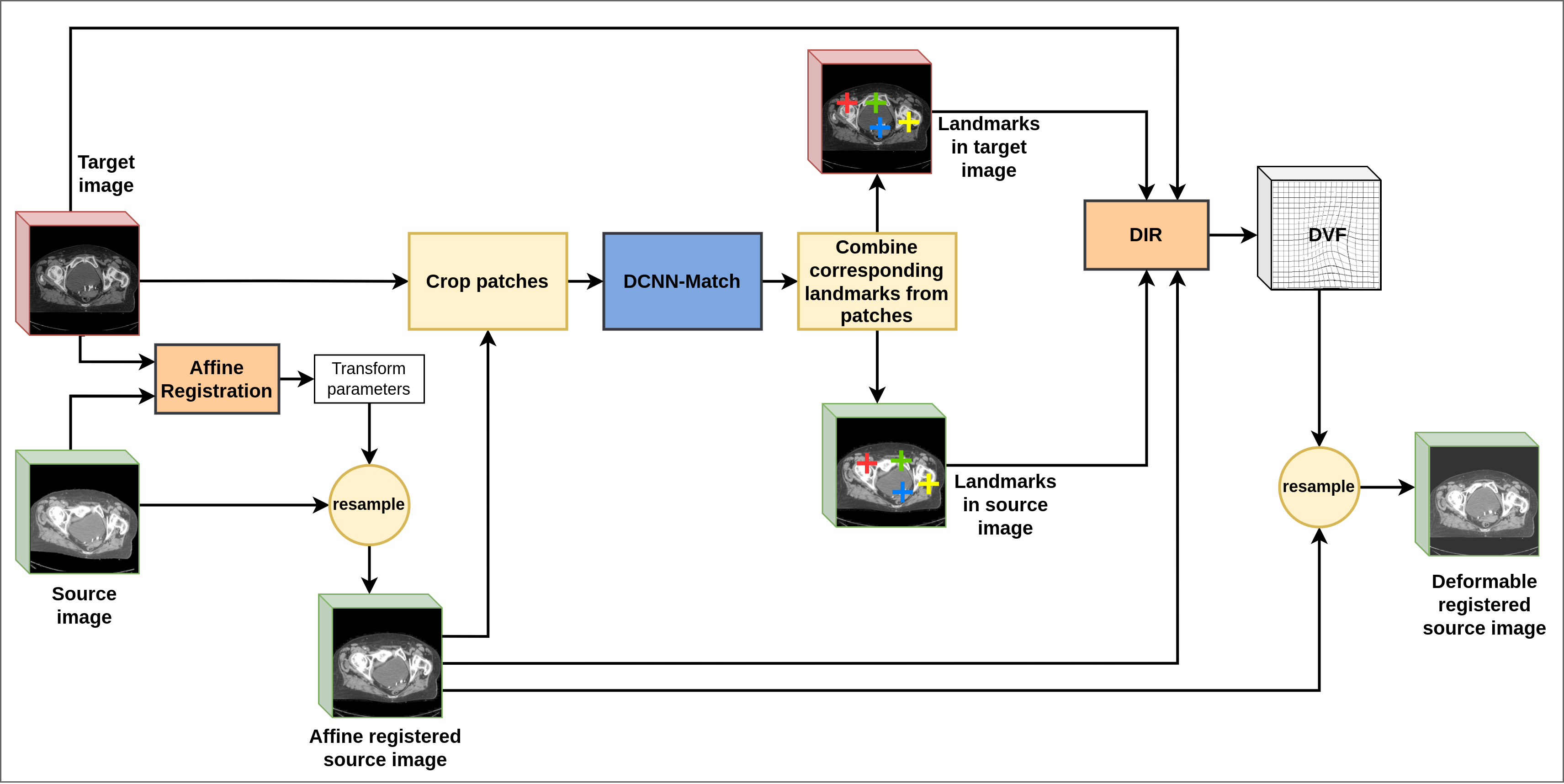}
    \caption{\textbf{DIR pipeline with automatic landmarks correspondence detection using DCNN-Match.} The source image is affine registered with the target image followed by automatic landmarks correspondence detection using DCNN-Match. DCNN-Match provides the locations of corresponding landmarks (shown with similar colored cross-hairs) in both the target and affine registered source image. The DIR module finds a DVF by utilizing the additional guidance information from automatic landmark correspondences. The final transformed (deformable registered) source image is obtained by resampling the affine registered source image according to the obtained DVF.}
    \label{fig:DIR pipeline}
\end{figure*}

We integrated DCNN-Match with the open-source registration software Elastix \cite{KleinStaring:2010, Elastix2:2014, SimpleElastix} to create a pipeline for DIR that utilizes the additional guidance information from automatic landmark correspondences. A schematic of the DIR pipeline is provided in Figure \ref{fig:DIR pipeline}.

DIR requires calculation of a DVF that maps each spatial location in the target image to a spatial location in the source image. In Elastix, the DVF is parameterized by B-splines and the coefficients of B-splines are optimized by non-linear optimization. We align the source CT scans with the target CT scans using affine registration before performing DIR. The parameters of the 3D affine transformation matrix (i.e., translation, rotation, scale, and shear) are optimized by maximizing the normalized mutual information between the target and source scans. The target and the affine registered source CT scan are input to the DCNN-Match, which provides the locations of corresponding landmarks in both the scans. The DIR module in Elastix takes the target image, affine registered source image, and the pairs of corresponding landmarks in both the images as input. The DIR is performed by optimizing the following objective function:

\begin{multline}\label{eq1}
    f_{Guidance} = weight_0 \,AdvancedMattesMutualInformation \\ + weight_1\, TransformBendingEnergyPenalty \\
    + weight_2\, {CorrespondingPointsEuclideanDistanceMetric}
\end{multline}
where $AdvancedMattesMutualInformation$ represents the maximization of mutual information between two scans (for details refer to \cite{MIPaper}), $TransformBendingEnergyPenalty$ is a regularization term that penalizes large transformations, and $CorrespondingPointsEuclideanDistanceMetric$ is used for minimizing the Euclidean distance between the landmarks in the target CT and the landmarks in the source CT. $weight_0$, $weight_1$, and $weight_2$ control the relative contribution of each term towards the objective function. 

\subsection{Implementation}
\label{sec:implementation}

The DIR pipeline was developed in Python. We used the PyTorch framework \cite{paszke2017automatic} for developing DCNN-Match. The training was done on an RTX 2080 Ti GPU and took approximately 21 hours. The weights of DCNN-Match were initialized using the He norm method \cite{he2015delving}. The training was done using the Adam optimizer \cite{adamConference} with a learning rate of $1e^{-4}$. The neural network weights were regularized by using a weight decay of $1e^{-4}$.

We randomly cropped 3D patches of dimension 128 $\times$ 128 $\times$ 48 from the entire CT scan volume and used them as target images. The source images were generated on-the-fly by applying one of the following random transformations on the target images: translation, rotation, scale, or elastic transformations. The magnitudes of the affine transformations along all axes were sampled from the following uniform distributions: $U(-12 mm, 12 mm)$, $U(-20^{\circ}, 20^{\circ})$, and $U(0.9, 1.1)$ for translation, rotation, and scale respectively. The elastic transformations were applied so as to simulate the two types of soft tissue deformations present in the lower abdominal scans: a) large local deformations e.g., bladder filling, b) small tissue deformations everywhere in the image. The large local deformations were simulated by a 3D Gaussian DVF ($DVF_{large}$) of magnitude at center $= U(2 mm, 24 mm)$ and $\sigma = U(64 mm, 128 mm)$ at a random location in the image. The small deformations everywhere in the image were simulated by Gaussian smoothing of a random DVF ($DVF_{small} = U(1 mm, 12 mm)$) at each location. $DVF_{large}$ and $DVF_{small}$ were additively applied to the target image to generate the source image with elastic transformation.

\subsection{Hyperparameters}
\label{sec:hyperparameters}
 
 Apart from the conventional hyperparameters involved in designing and training a DCNN e.g., network depth and width, optimizer, and learning rate, there are two hyperparameters specific to DCNN-Match: patch dimensions and the number of sampling points during training ($K$). As indicated in the previous section, we used a patch size of 128 $\times$ 128 $\times$ 48 (256 mm $\times$ 256 mm $\times$ 96 mm). This way the neural network's Field-Of-View (FOV) was maximum given the network depth and GPU memory constraints, which ensured that the landmark correspondences could be learned for deformations as large as 128 mm in-plane and 48 mm along the transverse axis. Similar to \cite{GrewalLandmark}, $K = 512$ was used based on the visual inspection that the predicted landmarks in the validation set (details in section \ref{subsection:train-val-set}) covered the image sufficiently. 
 
In Elastix, we used the advanced mattes mutual information as a similarity metric because it has been found successful in earlier studies on DIR \cite{ghose2015review}. For deciding other hyperparameters such as the number of iterations, step size, step decay, $weight_0$, $weight_1$, and $weight_2$, we used the development set (details in section \ref{subsection:train-val-set}). For this purpose, the pairs of target and source images were generated in a manner similar to the training set. 100 locations were sampled randomly on the target image and their corresponding location in the source image was established by transforming the coordinates with the inverse DVF used for generating the source image. The hyperparameters were tuned based on the following observations on the development set: the transformed source image after registration was not distorted and showed no visible folding, the image alignment at the 100 randomly sampled locations improved after registration. The exact configuration of Elastix used for affine registration and DIR is provided in the Appendix \ref{appendix: elastix parameter maps}.

\subsection{Data}
\label{sec:data}

\begin{figure}[h!]
    \centering
    \includegraphics[width=\linewidth]{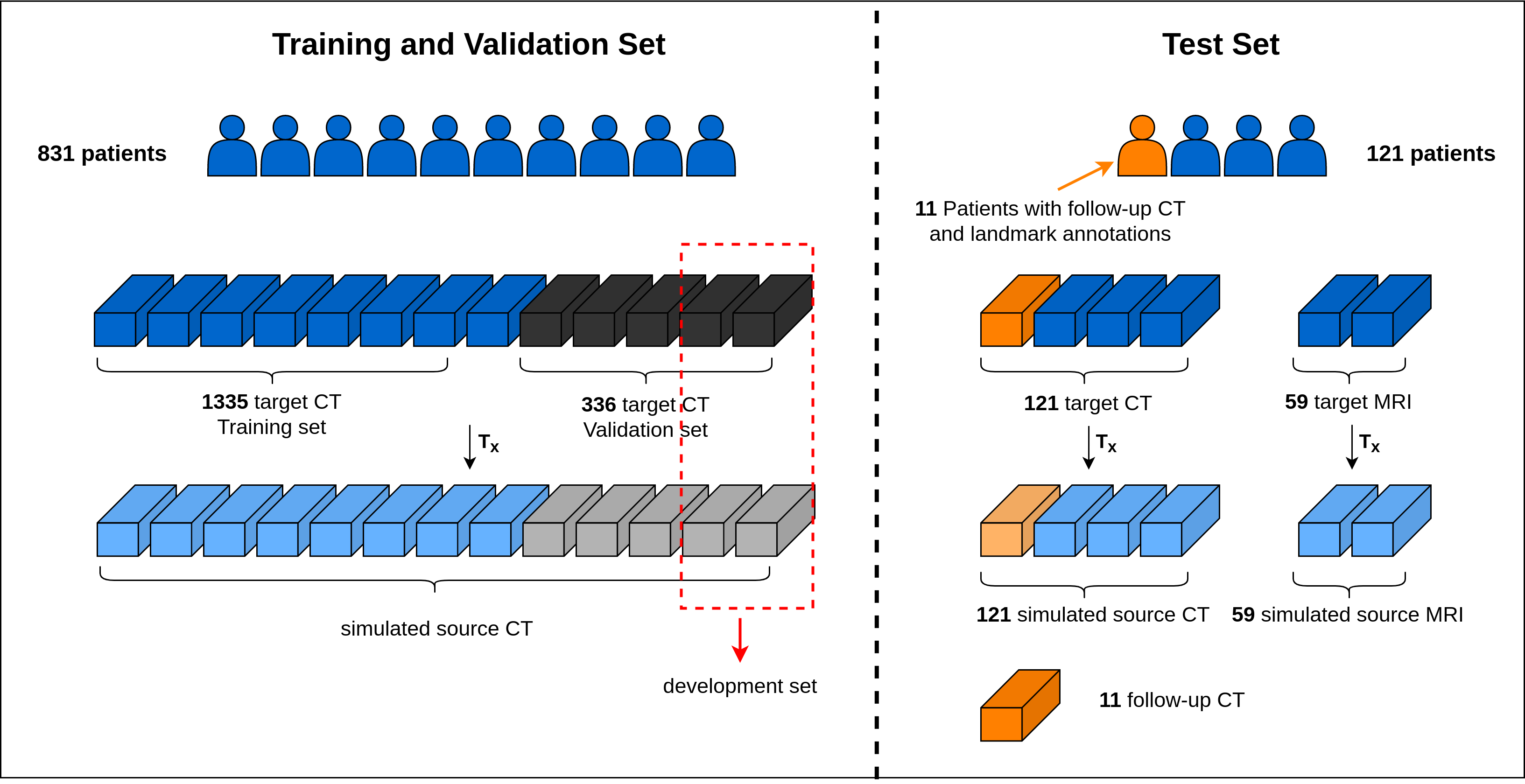}
    \caption{\textbf{Data Overview.} The vertical dashed gray line depicts the patient-level split between the training and validation set, and test set.}
    \label{fig:data}
\end{figure}

An overview of the data is provided in Figure \ref{fig:data}. We retrospectively included the CT and MRI scans from female patients (age range 22 - 95 years), who received radiation treatment in the lower abdominal region between the year 2009 and 2019 at the Amsterdam University Medical Centers, location AMC, the Netherlands. The data was transferred in anonymized form through a data transfer agreement. A subset of these scans was the same as used in a previous study \cite{GrewalLandmark}.

\subsubsection{Training and validation set}
\label{subsection:train-val-set}

A total of 1671 CT scans of 831 patients were used for developing the approach: 1335 CT scans for training and 336 CT scans for validation. A subset containing 10 CT scans from the validation set (referred to as the development set) was used to tune the hyperparameters of the DIR pipeline. All the CT scans were resampled to have 2 mm $\times$ 2 mm $\times$ 2 mm voxel spacing and the image intensities were converted from the Hounsfield units to a range of 0 to 1 after windowing.

\begin{SCfigure}
    \centering
    \includegraphics[width=0.7\textwidth]{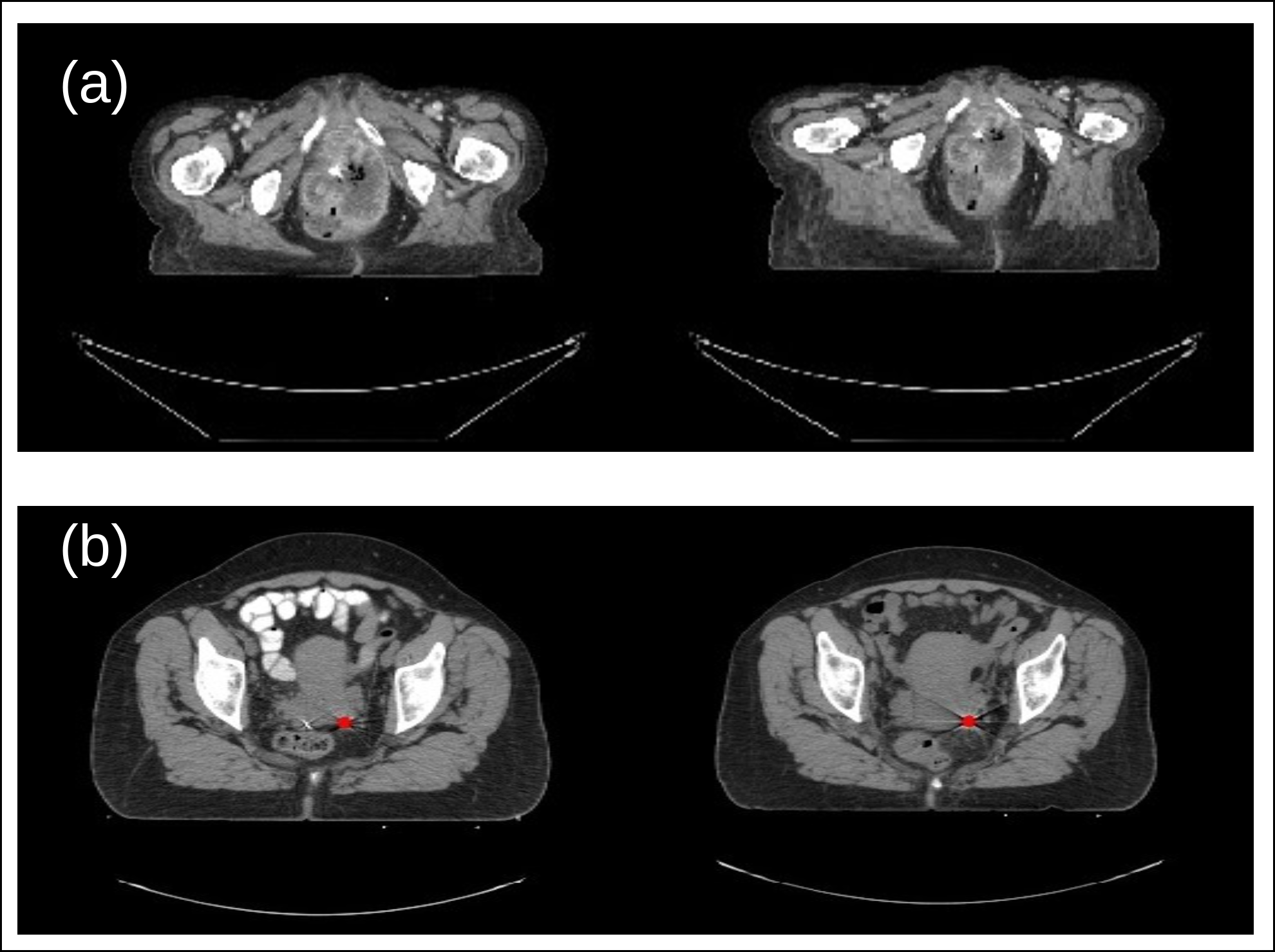}
    \caption{\textbf{Transverse slices from representative examples.} (a) simulated deformations test set: the source CT (right) is obtained by applying an elastic  transformation to the target CT (left). (b) clinical deformations test set: the landmark at the location of a fiducial marker (shown with red dot) in the target (left) and source (right) CT is shown. Note the appearance difference in the bowel due to contrast.}
    \label{fig:contrast example}
\end{SCfigure}

\subsubsection{Simulated deformations test set - CT}
\label{subsection:simulated test set - CT}

We tested the performance of DCNN-Match and the DIR pipeline on a curated dataset of 121 CT scans belonging to 121 patients, who received radiation treatment for cervical cancer. The mean FOV of acquisition of the CT scans was 546 mm $\times$ 546 mm $\times$ 368 mm and the scans were resampled to 2 mm $\times$ 2 mm $\times$ 2 mm voxel spacing. The available CT scans were used as target images and corresponding source images were simulated by applying random elastic transformations to the target CT scans according to the method described in the section \ref{sec:implementation} above. Further, an example of the simulated deformation and the obtained source CT is shown in Figure \ref{fig:contrast example} (a).

In each pair of target and source image, 100 corresponding locations were sampled with uniform random distribution. These sampled locations were used as validation landmarks for assessing the performance of DCNN-Match and the DIR pipeline.

\subsubsection{Clinical deformations test set - CT}
\label{subsection:clinical test set - CT}

The CT scans in a clinical setup exhibit complex bio-mechanical deformations including discontinuities in the deformation field around sliding tissues and large deformations that may not be Gaussian. The random Gaussian DVF used for deforming the images to obtain a simulated test set is an oversimplification of the underlying situation. Therefore, it is essential to investigate if the observations on the simulated deformations test set hold in the clinical setting as well. To this end, additional CT scans (referred to as follow-up scans) were searched in the clinical database for a subset of patients in the test set (11 patients). The first CT scans from these patients were used as target images and the corresponding follow-up CT scans were used as source images.

Corresponding landmarks at 29 locations were manually identified in each target and source CT scan by a clinical expert. These landmarks included six fiducial markers in the vaginal wall, and anatomical landmarks e.g., aortic bifurcation, cervical os, and os coccygis. Since clinically available scans were used, the number of fiducial markers were different in each patient in accordance with the treatments given to the patients. The majority of the patients' scans had three fiducial markers, while some had less or more. If a patient's scan had less than three fiducial markers, calcification (if present) in corresponding anatomical locations were used as landmarks. If a patient's scan had more than three fiducial markers, only three of them were used. An example landmark location is shown in Figure \ref{fig:contrast example} (b) and the complete list of landmark locations is provided in Appendix \ref{appendix: landmarks list}.

\subsubsection{Simulated deformations test set - MRI}
\label{subsection:simulated test set - MRI}

MRI scans of 59 cervical cancer patients (subset of the 121 cervical cancer patients mentioned in section \ref{subsection:simulated test set - CT}, who received brachytherapy treatment) acquired during brachytherapy treatment delivery were used to investigate the generalization capability of DCNN-Match. The mean FOV of acquisition of the MRI scans was 199 mm $\times$ 199 mm $\times$ 152 mm and the scans were resampled to 2 mm $\times$ 2 mm $\times$ 2 mm voxel spacing. The pairs of source and target scans were generated in a similar way to the CT scans (section \ref{subsection:simulated test set - CT}). 

\subsection{Experiments}
\label{section:experiments}

We conducted three types of experiments. The first type of experiments were aimed to gain insights in the working of DCNN-Match by changing the $DescriptorMatchingLoss$ (sections \ref{subsection:exp_descriptor_loss} and \ref{subsection:exp_margin_in_hinge_loss}). The second type of experiments were done to investigate the effect of automatically predicted landmark correspondences on the performance of DIR (section \ref{subsection:exp_additional_guidance_effect}). We also investigated how the changes in $DescriptorMatchingLoss$ affected the added value of the automatic landmark correspondences toward the performance of DIR. Third, we investigated the generalization capability of DCNN-Match on a different modality (section \ref{subsection:generalization_mri}).

\subsubsection{Descriptor Loss}
\label{subsection:exp_descriptor_loss}

We trained three versions of DCNN-Match, each with a different $DescriptorMatchingLoss$. The first version was trained with only the $DescriptorHingeLoss$ defined in Equation \eqref{eq:hingeloss}. This version is referred to as DCNN-Match Hinge.  DCNN-Match Hinge was trained with $m_{pos} = 0$ and $m_{neg} = 1$. In the second version, only $DescriptorCELoss$ Equation \eqref{eq:CE} was employed. We refer to this version as DCNN-Match CE. Next, we trained the network with a linear combination of $DescriptorHingeLoss$ and $DescriptorCELoss$ Equation \eqref{eq:DescriptorMatchingLoss}, which is referred to as DCNN-Match Hinge+CE.

\subsubsection{Positive Margin in the Hinge Loss}
\label{subsection:exp_margin_in_hinge_loss}

We considered that the L2-norm of the descriptor pairs of highly deformed regions would be high and these pairs would be difficult to match. Further, it is intuitive to think that the landmark matches in regions of high deformation would provide more added value to the DIR approach. To allow the network to focus more on matching these pairs, we trained DCNN-Match Hinge+CE with two values for $m_{pos}$: 0.1 and 0.2. These versions are referred to as DCNN-Match Hinge0.1+CE and DCNN-Match Hinge0.2+CE, respectively. The value of $m_{pos} > 0$ in the $DescriptorHingeLoss$ makes the loss term 0 for descriptor pairs whose L2-norm is less than $m_{pos}$ i.e., the network already identifies the descriptor pairs as matching. Thus, the gradients are influenced only by the descriptor pairs which are difficult to match. Consequently, the network should be able to predict difficult landmark correspondences in the highly deformed regions accurately.

\subsubsection{Effect of Additional Guidance from Automatic Landmark Correspondences}
\label{subsection:exp_additional_guidance_effect}

To assess the effect of additional guidance from automatic landmark correspondences on the DIR, we compared the results from the DIR pipeline with ($weight_2 = 0.01$ in Equation \eqref{eq1} as obtained from hyperparameter tuning on the development set) and without ($weight_2 = 0$ in Equation \eqref{eq1}) automatic landmarks correspondence detection.

\subsubsection{Generalization to MRI dataset}
\label{subsection:generalization_mri}

Given the capability of deep neural networks to learn robust features, and the self-supervised nature of our training approach, optimistically one would expect that the developed approach would generalize to different datasets. To this end, we tested DCNN-Match on pairs of MRI scans containing simulated deformations (described in section \ref{subsection:simulated test set - MRI}) without retraining. Compared to the training set, the MRI scans were not only different in imaging modality, but also in the FOV of acquisition.

\subsection{Evaluation}
\label{sec:evaluation}
\subsubsection{Spatial Matching Errors of Landmark Correspondences}
\label{subsection:evaluation -> spatial matching errors}

In the simulated deformations test set, the landmarks on the source CT scans were projected on the target CT scans using the known transformation between them. The Euclidean distances between the landmarks on the target CT scans and the projection of their corresponding landmarks predicted by the network were calculated. The Euclidean distance gives a measure of the spatial matching error of the predicted landmark correspondences. The spatial matching errors were compared between all versions of DCNN-Match.

Quantitative analysis of the spatial matching errors of the predicted landmark correspondences is not feasible in the clinical deformations test set due to the absence of the ground truth DVF. To provide some insights into the performance on the clinical deformations test set, we conducted a  validation study on a subset of the data. For this purpose, we randomly sampled 75 predicted landmarks from DCNN-Match CE in two patients (total 150 landmark correspondences). A radiation oncologist (henceforth, referred to as clinician) ranked these landmark correspondences on a 3 point Likert scale: 1 = good match (roughly within 5 mm distance), 2 = moderate match (roughly within 10 mm distance), 3 = poor or wrong match (roughly more than 15 mm distance) in a 3D (axial, sagittal, and coronal) image viewer. The (approximate) spatial matching errors were calculated based on the ranking provided by the clinician. The clinician also labelled the anatomical location of the landmarks in target CT scans according to the following categories: a) bony anatomy, b) soft tissue (i.e., muscles, fatty tissue, and fascia), c) bowel i.e., rectum, large and small bowel including gas pockets, d) other (including organs and blood vessels i.e., veins and arteries). We also analyzed the spatial matching errors of the predicted landmark correspondences separately for each anatomical category.

\subsubsection{Target Registration Error}
\label{subsection:evaluation->TRE}

In the clinical deformations test set, we transformed the manually annotated landmarks in the target images according to the estimated DVF after DIR using the transformix module in SimpleElastix \cite{SimpleElastix} (documentation on using transformix in SimpleElastix can be found at \href{https://simpleelastix.readthedocs.io/PointBasedRegistration.html}{SimpleElastix documentation} and \href{https://elastix.lumc.nl/download/elastix-5.0.1-manual.pdf}{Elastix manual}). We calculated their Euclidean distance with the corresponding landmarks in the source image. This measure is often referred to as ``Target Registration Error" or TRE. We calculated the TRE values after initial affine registration and before the DIR ($TRE_{before}$) and after DIR ($TRE_{after}$) for all experiments. In the simulated deformations test set, TRE calculations were done using the randomly sampled validation landmarks described in section \ref{sec:data}.

It should be noted that the TRE calculations were done in image space i.e., the landmarks were represented by the center of a voxel. We chose this setup because the automatic landmarks are predicted in image space. However, this setup may give rise to discretized TRE values.

\subsubsection{Landmark Correspondences vs. Underlying Deformation}
\label{subsection:evaluation -> spatial density}
 
It is intuitive to think that the DIR performance in a specific region is dependent on the underlying deformation in that region. Concordantly, the distribution of landmarks with respect to the underlying deformation would impact the additional guidance provided by the landmarks overall. Therefore, it is important to investigate the choice of landmark locations by the network with respect to the extent of deformation at those locations. To this end, we partitioned the voxels in the source images in the simulated deformations test set into bins of different deformations. For each DCNN-Match variant, the spatial density of predicted landmark correspondences was calculated in each bin of the underlying deformation by dividing the number of landmarks with the number of voxels in each bin.
 
Similarly, we calculated the percentage of automatic landmarks below 4 mm spatial matching errors (as a surrogate for landmarks correspondence accuracy) and TRE values of validation landmarks (as a measure of DIR performance) in each deformation region. The threshold of 4 mm was chosen because the same threshold was used during training. In each deformation region, we analyzed the TRE values in light of the spatial density and landmarks correspondence accuracy to gain insights about what aspects of automatic landmarks affect the DIR performance.

\subsubsection{Determinant of Spatial Jacobian}

Evaluating the performance of DIR is a difficult task and TRE can only give an estimate of performance on sparse image locations. Moreover, TRE can give a biased perspective of the DIR performance because of the observer subjectivity in the manual annotation of landmark locations. In order to assess whether the obtained DVF is anatomically plausible or not, the determinant of the spatial Jacobians of the DVF is a good measure. The negative values in the determinant of the spatial Jacobian represent singularities in the DVF and indicate image folding in those regions. Therefore, we also investigated the determinant of the spatial Jacobians of the obtained DVFs after DIR.

\subsection{Statistical Testing}
\label{sec:statistics}

The statistical testing was done using IBM SPSS Statistics for Ubuntu (Version 27.0, IBM Corp. Released 2020. Armonk, NY: IBM Corp)\cite{SPSS}. We tested the null hypothesis that the $TRE_{after}$ values in the test sets were the same in the following experimental scenarios: DIR without additional guidance from corresponding landmarks, and DIR with additional guidance from five different variants of DCNN-Match.

Kolmogorov-Smirnov tests for normality revealed that the $TRE_{after}$ values were not normally distributed in any of the experimental scenarios. Therefore, we used the related samples Friedman's two way Analysis of Variance by Ranks test followed by post-hoc pairwise comparisons using Dunn-Bonferroni test \cite{dunn1964multiple}. An alpha of 0.05 with Bonferroni correction for multiple comparisons was considered significant. 


\section{Results}
\label{section:results}
\begin{figure*}[h!]
    \centering
    \includegraphics[width=0.95\textwidth]{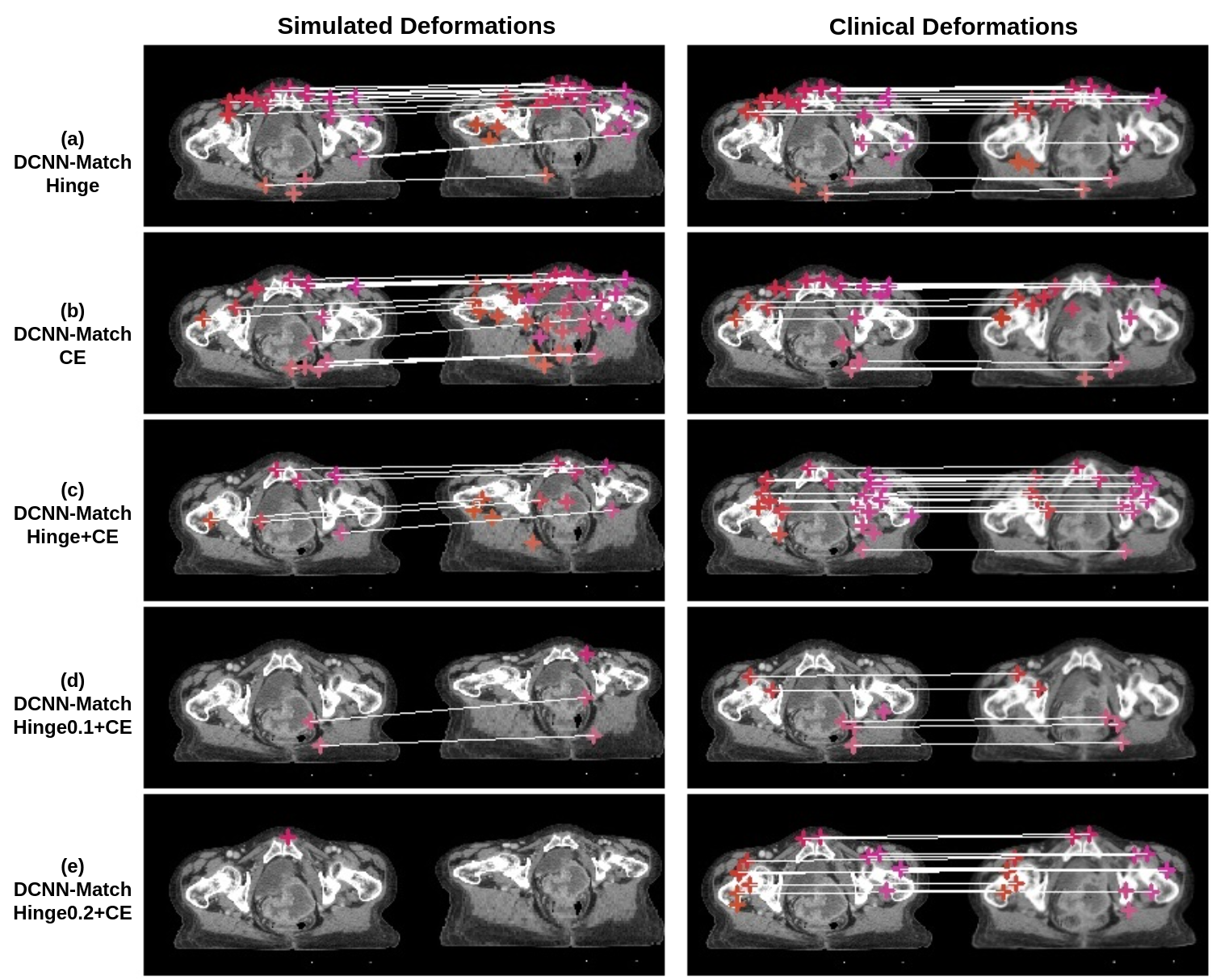}
    \caption{\textbf{Visualization of predicted landmark correspondences} by (a) DCNN-Match Hinge, (b) DCNN-Match CE, (c) DCNN-Match Hinge+CE, (d) DCNN-Match Hinge0.1+CE, and (e) DCNN-Match Hinge0.2+CE. A transverse slice from target and source CTs in the simulated deformations test set (left) and the clinical deformations test set (right) is shown. The corresponding landmarks are shown with the same colored cross-hairs in target and source image and a white line is drawn for in-slice corresponding landmarks. Please note that some corresponding landmarks may lie on a different slice and are therefore not visible in the figure.}
    \label{fig:number landmarks}
\end{figure*}

\renewcommand{\arraystretch}{1.5}
\begin{table*}[h!]
\caption{Number of predicted landmark correspondences per CT scan pair. Mean (M) $\pm$ standard deviation (SD), and range ($5^{th}$ percentile -- $95^{th}$ percentile) are provided.}
\label{tab:number of landmarks}
\begin{tabular}{m{1.6cm} m{2.45cm} m{2.45cm} m{2.45cm} m{2.45cm} m{2.45cm}}
\toprule
& DCNN-Match Hinge & DCNN-Match CE & DCNN-Match Hinge+CE & DCNN-Match Hinge0.1+CE & DCNN-Match Hinge0.2+CE \\ \midrule
\rowcolor[HTML]{C0C0C0}
\multicolumn{6}{c}{\textbf{Simulated Deformations}}\\[1.0ex]
M $\pm$ SD & 5488 $\pm$ 2258 & 7761 $\pm$ 2540 & 1698 $\pm$ 888 & 1735 $\pm$ 959 & 1220 $\pm$ 871\\
Range & 2160 -- 9580 & 2999 -- 11400 & 595 -- 3462 & 563 -- 3563 & 244 -- 3028\\ [2.0ex]
\rowcolor[HTML]{C0C0C0}
\multicolumn{6}{c}{\textbf{Clinical Deformations}}\\[1.0ex]
M $\pm$ SD & 3708 $\pm$ 1052 & 7427 $\pm$ 1682 & 946 $\pm$ 391 & 1062 $\pm$ 479 & 511 $\pm$ 307\\
Range & 2563 -- 5344 & 5394 -- 10340 & 491 -- 1569 & 455 -- 1819 & 193 -- 1000\\ \bottomrule
\end{tabular}
\end{table*}

The average inference time of DCNN-Match variants for predicting landmark correspondences in one CT scan pair was $\sim$20s. A representative example of predicted landmark correspondences is shown in Figure \ref{fig:number landmarks}. The images in the figure are shown with the couch table cropped for better visualization, but the automatic landmark correspondence detection as well as DIR were performed on full CT scans without any cropping.

\begin{figure}
    \centering
    \includegraphics[width=\textwidth]{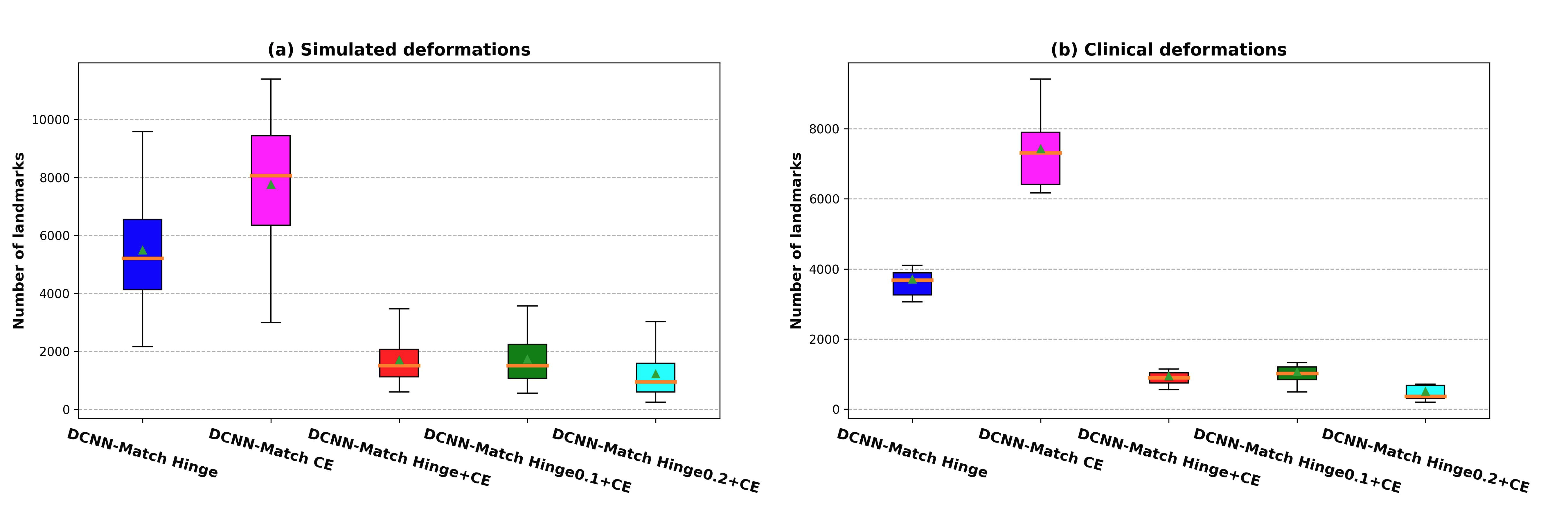}
    \caption{\textbf{Distribution of predicted automatic landmark correspondences across patients.} The boxes extend from the lower to upper quartile values of the data, with a line at the median. Mean is shown with a triangular marker and whiskers represent the range from $5^{th}$ to $95^{th}$ percentile.}
    \label{fig:boxplot number}
\end{figure}

\subsection{Number of Landmark Correspondences}

The number of landmark correspondences predicted per image on the simulated test set and clinical test set is described in Table \ref{tab:number of landmarks} and Figure \ref{fig:boxplot number}. As can be seen in Table \ref{tab:number of landmarks} and Figure \ref{fig:boxplot number}, DCNN-Match Hinge and DCNN-Match CE approaches predicted a large number of landmarks per CT scan pair. In DCNN-Match Hinge+CE, the use of an auxiliary loss allows for applying an additional constraint on the landmark correspondences. Consequently, the number of predicted landmark correspondences per image was fewer than with using either of the loss separately. Further, the DCNN-Match Hinge0.1+CE and DCNN-Match Hinge0.2+CE predicted even fewer landmarks per CT scan pair, possibly due to the additional constraint posed by the positive margin $m_{pos}$ used in the Hinge loss. It should be noted that irrespective of the differences within different DCNN-Match variants, a considerable number of landmark correspondences were predicted by all of them in both the simulated as well as the clinical deformations test set.

\subsection{Spatial Matching Errors of Landmark Correspondences}
\begin{SCfigure}
    \centering
    \includegraphics[width=0.7\textwidth]{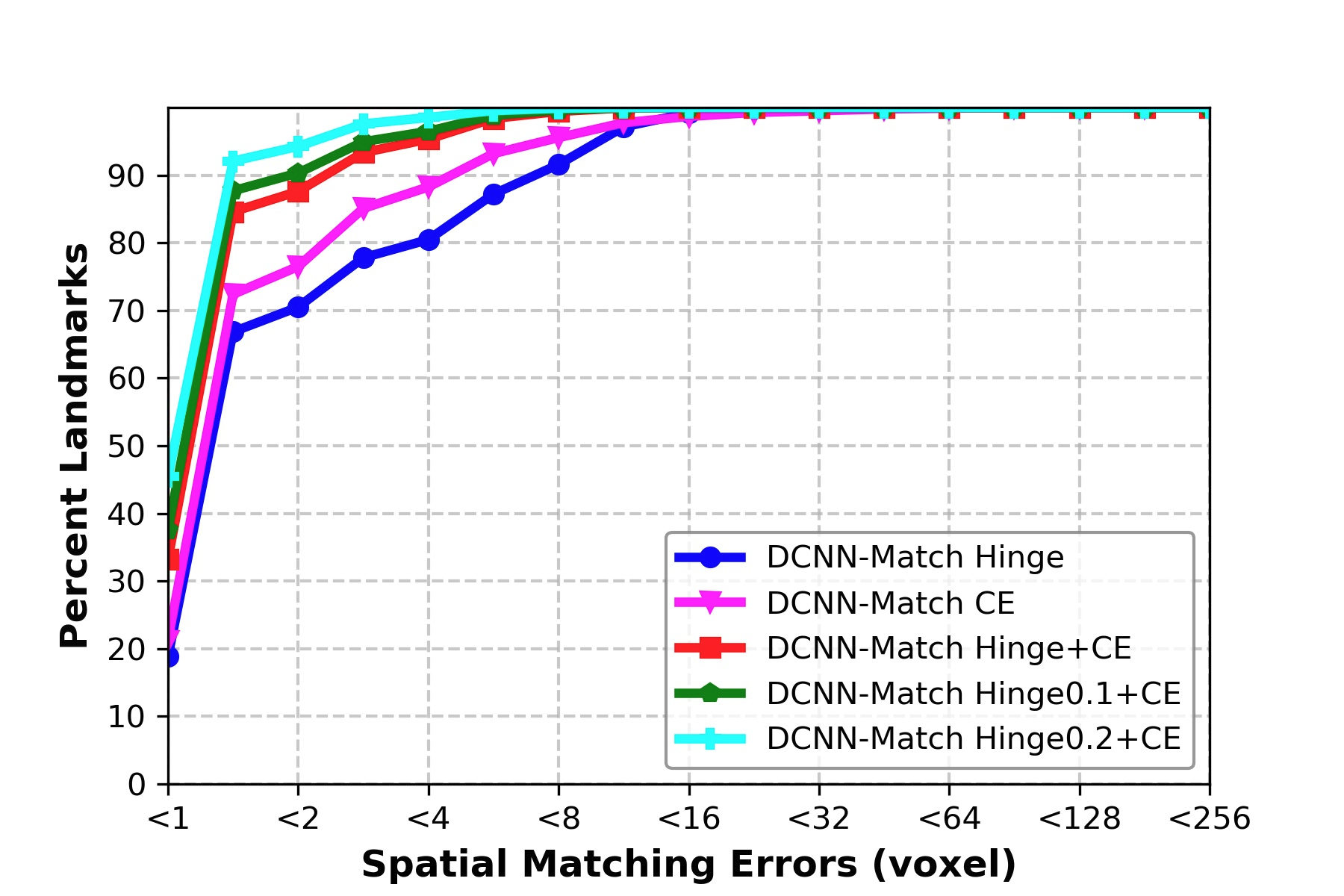}
    \caption{Cumulative distribution of the landmarks with respect to the spatial matching errors for different versions of DCNN-Match on the simulated deformations test set - CT.}
    \label{fig:spatial matching errors}
\end{SCfigure}

\begin{figure}
    \centering
    \includegraphics[width=\textwidth]{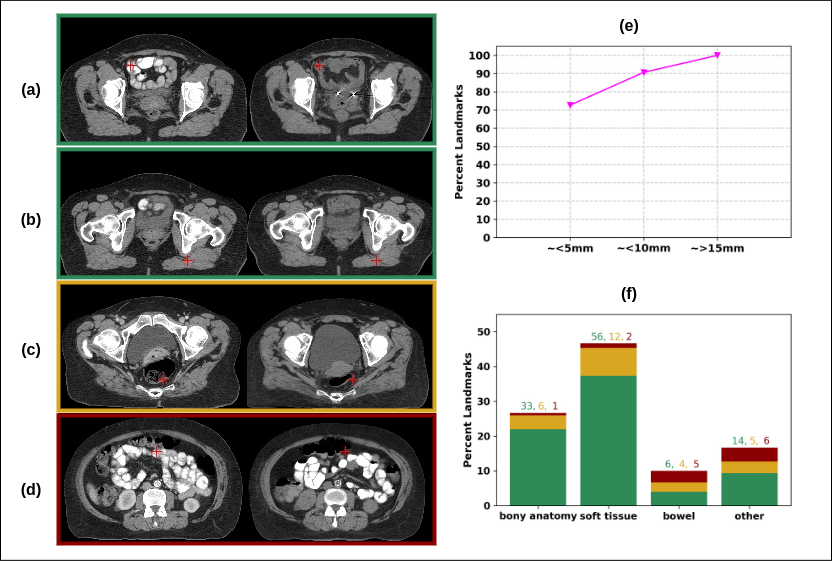}
    \caption{\textbf{Validation of landmark correspondences in clinical deformations test set.} Representative examples of (a)-(b): good match despite contrast variation and difference in muscle deformation, (c): moderate match, and (d): wrong match. (e): Cumulative distribution of landmarks with respect to (approximate) spatial matching errors. (f): Distribution of landmarks in different anatomical categories. The bars are shaded in proportion to the number of landmarks corresponding to a rank: green = good, yellow = moderate, red = wrong. In each anatomical category, the total number of landmarks representing a rank is provided in the text above bars in the corresponding color.}
    \label{fig:clinical validation}
\end{figure}

\begin{figure}
    \centering
    \includegraphics[width=\textwidth]{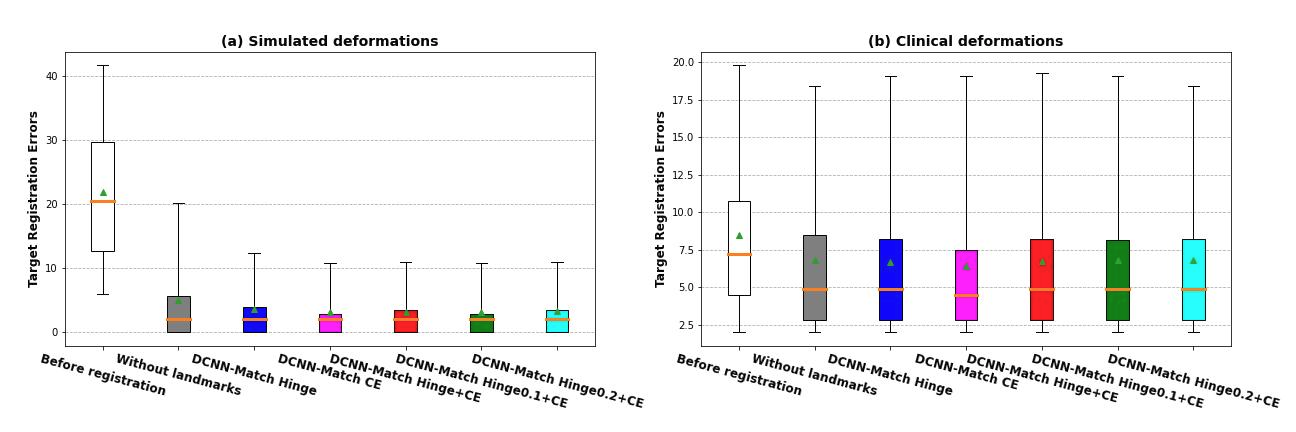}
    \caption{\textbf{Distribution of Target Registration Errors (TRE).} The boxes extend from the lower to upper quartile values of the data, with a line at the median. Mean is shown with a triangular marker and whiskers represent the range from $5^{th}$ to $95^{th}$ percentile.}
    \label{fig:boxplot TRE}
\end{figure}
The cumulative distribution of the predicted landmark correspondences in the simulated test set is plotted against their spatial matching errors in Figure \ref{fig:spatial matching errors}. Both DCNN-Match Hinge and DCNN-Match CE predicted more than 70\% landmarks with less than 2 voxels (equivalent to 4 mm) spatial matching error. But, DCNN-Match CE predicted a higher percentage of landmarks within a specific spatial matching error as compared to DCNN-Match Hinge. The decrease in spatial matching errors could be attributed to the added parameters used in the dedicated descriptor matching module in DCNN-Match CE as opposed to the parameter-free module in DCNN-Match Hinge. Further, DCNN-Match Hinge+CE takes advantage of the auxiliary loss and therefore, the landmark correspondences are predicted with lower spatial matching errors. About 90\% of the predicted landmarks had a spatial matching error of less than 4 mm.

As expected, training with $m_{pos}>0$ yielded landmarks with lower spatial matching errors as compared to DCNN-Match Hinge+CE (Figure \ref{fig:spatial matching errors}). Specifically for DCNN-Match Hinge0.2+CE, more than 90\% of the predicted landmark correspondences had spatial matching errors of less than 1 voxel, which is equivalent to 2 mm (image resolution). This finding indicates the potential of the automatic landmark correspondences predicted by the DCNN-Match variant for use in clinical applications.

\subsection{Spatial Matching Errors in Clinical Data}
In Figure~\ref{fig:number landmarks}, the predicted landmark correspondences from DCNN-Match variants on a representative transverse slice from the clinical deformations test set are shown for the reader's perusal. More examples are shown in Figure~\ref{fig:clinical validation} (a)-(d). The border colors indicate the ranking given by the clinician: green = good, yellow = moderate, red = wrong match. Figure~\ref{fig:clinical validation} (a) demonstrates a good match in the small bowel despite the difference of the underlying contrast and Figure~\ref{fig:clinical validation} (b) demonstrates a good match in the muscle despite a change in the muscle deformation. In Figure~\ref{fig:clinical validation} (c), both the landmarks are present in the rectum, but in different locations, although it was challenging to review because of the presence of the gas pocket and change in the muscle deformation. Figure~\ref{fig:clinical validation} (d) shows an example of a wrong match in the bowel. It is important to note the underlying challenges visible between the two scans in Figure~\ref{fig:clinical validation} (d) e.g., difference in contrast, and content mismatch due to presence of gas pockets.

Figure~\ref{fig:clinical validation} (e) shows that more than 72\% landmark correspondences were ranked as good match i.e., approximately within 5 mm distance and about 90\% landmark correspondences were ranked to be within 10 mm distance. These results indicate only a small performance difference in comparison to the simulated deformations test set (magenta curve in Figure~\ref{fig:spatial matching errors} and Figure~\ref{fig:clinical validation} (e)), which is expected due to the presence of additional challenges in the clinical data.  

Further, in Figure~\ref{fig:clinical validation} (f), the percentage of landmarks in bony anatomy, soft tissue, bowel, and other regions is plotted. The bars in the plot are shaded in green, yellow, and red colors in proportion to the ranking of the landmarks (green = good, yellow = moderate, red = wrong) in that anatomical category. It is worth noting that the wrong matches are mainly in the bowel region, where content mismatch may happen along with large deformations and intensity variations.

\subsection{Effect of Landmark Correspondences on DIR}

In Table \ref{tab:tre table}, the TRE values in the simulated and clinical deformations test sets are provided. In Figure~\ref{fig:boxplot TRE}, boxplots of TRE values are provided. In both test sets, there was a significant main effect of the experimental scenario (i.e., DIR without landmarks and with landmarks predicted by either one of the DCNN-Match variants) on the observed $TRE_{after}$ values, ($\chi$(5) = 6620.117, p = $0e^0$) in the simulated test set and ($\chi$(5) = 34.051, p = $0.000002$) in the clinical test set. Note that in the simulated test set, the sample size was quite large (100 landmarks per scan $\times$ 121 scans = 12100) giving rise to near zero p values in the statistical testing.

In the simulated deformations test set, the post-hoc comparisons revealed that $TRE_{after}$ values from registration using additional guidance from landmark correspondences predicted by any of the DCNN-Match variants were significantly lower than $TRE_{after}$ values from registration without using additional guidance from landmark correspondences. However, the strongest effect was observed with landmark correspondences from DCNN-Match CE (p = $0e^0$). 

On the clinical deformations test set, although the $TRE_{after}$ values from registration with the use of additional guidance by automatic landmark correspondences were smaller than the $TRE_{after}$ values from registration without using additional guidance from landmark correspondences, the differences were small.
Only the post-hoc comparison between $TRE_{after}$ values from registration by using landmark correspondences predicted by DCNN-Match CE and $TRE_{after}$ values from registration without using landmark correspondences yielded statistical significance after correction for multiple comparisons (p = $0.030$).

\renewcommand{\arraystretch}{1.5}
\begin{table*}[h!]
\centering
\caption{Target Registration Errors (TREs) in mm of pre-specified landmarks (for details refer to \ref{subsection:evaluation->TRE}) before DIR but after affine registration ($TRE_{before}$) and after DIR with different approaches ($TRE_{after}$). Mean (M) $\pm$ standard deviation (SD), and range ($5^{th}$ percentile -- $95^{th}$ percentile) are provided. Best TRE values are highlighted in bold. $^\ast$ represents significance in post-hoc comparison against $TRE_{after}$ without landmarks.}
\label{tab:tre table}
\begin{tabular}{m{1.8cm} m{2.5cm} m{2.4cm} m{2.4cm} m{2.4cm} m{2.4cm}}
\toprule
& & \multicolumn{2}{l}{\textbf{Simulated Deformations}} &  \multicolumn{2}{l}{\textbf{Clinical Deformations}}\\ \midrule
\rowcolor[HTML]{C0C0C0}
& & M $\pm$ SD & Range & M $\pm$ SD & Range \\[1.0ex]
\multicolumn{2}{l}{${TRE}_{before}$} & 21.99 $\pm$ 12.67 & 6.00 -- 41.76
& 8.50 $\pm$ 5.81 & 2.00 -- 19.96 \\\cmidrule{1-6}

\multirow{6}{*}{${TRE}_{after}$} 
& Without \newline landmarks & 5.07 $\pm$ 9.98 & 0.00 -- 20.20
& 6.85 $\pm$ 5.79 & 2.00 -- 19.12\\

& DCNN-Match Hinge & 3.58 $\pm$ 8.80 $^\ast$ & 0.00 -- 12.33
& 6.69 $\pm$ 5.84 & 2.00 -- 19.53\\

& DCNN-Match CE & \textbf{3.14 $\pm$ 8.61$^\ast$} & 0.00 -- 10.77
& \textbf{6.42 $\pm$ 5.79}$^\ast$ & 2.00 -- 19.94\\

& DCNN-Match Hinge+CE & 3.21 $\pm$ 8.63$^\ast$ & 0.00 -- 10.95
& 6.74 $\pm$ 5.77 & 2.00 -- 19.47\\

& DCNN-Match Hinge0.1+CE & 3.18 $\pm$ 8.62$^\ast$ & 0.00 -- 10.77
& 6.79 $\pm$ 5.83  & 2.00 -- 19.31\\

& DCNN-Match Hinge0.2+CE & 3.27 $\pm$ 8.65$^\ast$ & 0.00 -- 10.95
& 6.82 $\pm$ 5.86 & 2.00 -- 19.53\\
\bottomrule
\end{tabular}
\end{table*}

\subsection{Differential Effect of DCNN-Match variants on DIR}
\begin{figure*}[ht]
    \centering
    \includegraphics[width=\textwidth]{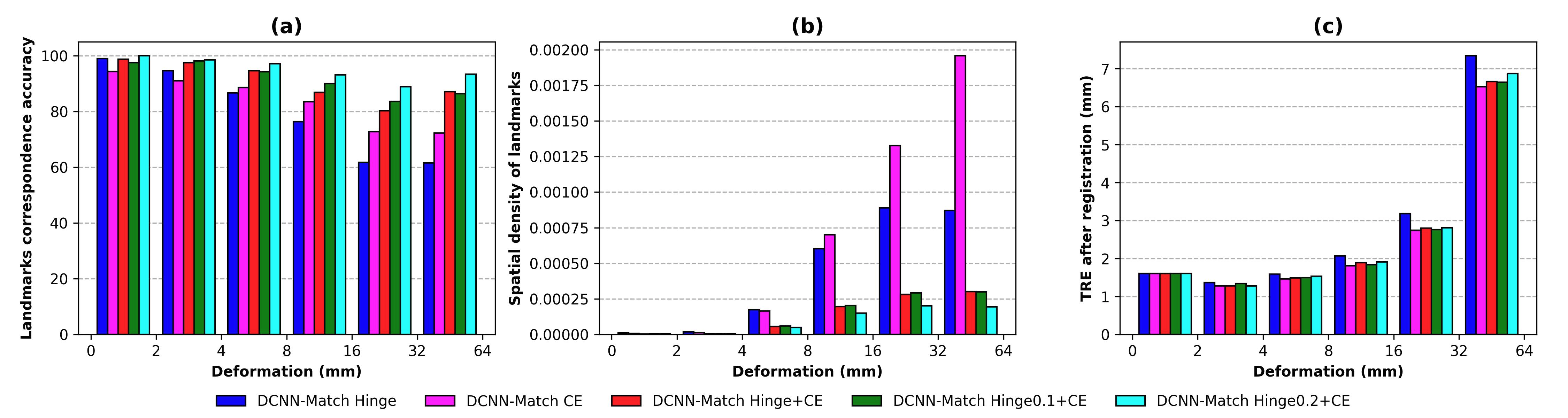}
    \caption{\textbf{Analysis of relation between aspects of landmark correspondences and DIR performance.} (a) Landmarks correspondence accuracy in different regions of underlying deformation represented by the percentage of landmarks predicted within 4 mm spatial matching errors. (b) Spatial density of landmarks (number of landmarks per voxel) predicted in different regions of underlying deformation. (c) TRE of validation landmarks after DIR by using automatic landmarks.}
    \label{fig:landmarks vs deformation}
\end{figure*}

The post-hoc analysis indicated that the landmarks predicted by DCNN-Match CE had significantly more added value (as reflected by the $TRE_{after}$ values) as compared to the landmarks predicted by DCNN-Match Hinge on the simulated test set (p = $0e^0$). However, a similar finding could not be corroborated on the clinical deformations test set -- pairwise comparison of $TRE_{after}$ values obtained by DCNN-Match CE and DCNN-Match Hinge did not yield significance after correcting for multiple comparisons (p = $0.406$).

Based on the observed spatial matching errors, it is intuitive to expect that DCNN-Match Hinge+CE would yield lower TRE values after registration as compared to DCNN-Match CE. However, surprisingly this is not the case (Table \ref{tab:tre table}). $TRE_{after}$ values using DCNN-Match CE were significantly lower than $TRE_{after}$ values using DCNN-Match Hinge+CE in the simulated deformations test set (p = $0.013$). In the clinical deformations test set also, the $TRE_{after}$ values using DCNN-Match CE were significantly lower than $TRE_{after}$ values using DCNN-Match Hinge+CE (p = $0.046$).

Furthermore, the TRE values after registration were not affected by increasing $m_{pos}$ in the simulated test set. The post-hoc pairwise comparisons of $TRE_{after}$ values by using DCNN-Match Hinge+CE vs DCNN-Match Hinge0.1+CE were not significant (p = 0.783) on the simulated deformations test set. In fact, the $TRE_{after}$ values by using DCNN-Match Hinge0.2+CE values were significantly higher than $TRE_{after}$ values by using DCNN-Match Hinge0.1+CE (p = $0.000244$). This indicates that even though an increase in $m_{pos}$ predicts landmark correspondences with lower spatial matching errors, there is no additional benefit toward DIR performance. The observations on clinical deformations also corroborated the findings on simulated deformations. None of the post-hoc comparisons between experimental scenarios with different $m_{pos}$ values were significantly different in the clinical deformations test set.

Overall, the results from pairwise comparisons between the $TRE_{after}$ indicate that the added value of the automatic landmark correspondences towards the improvement of DIR performance is dependent on the underlying approach for identifying automatic landmark correspondences.

\subsection{Relation between Aspects of Automatic Landmarks and DIR Performance}

In figure \ref{fig:landmarks vs deformation} (a), the landmarks correspondence accuracy (averaged over 121 patients) as described in section \ref{subsection:evaluation -> spatial density} in the regions of different underlying deformation is plotted for each DCNN-Match variant. As can be seen, the correspondence accuracy of the automatic landmarks predicted by DCNN-Match Hinge deteriorated as the underlying deformation increased. A similar trend was observed for DCNN-Match CE, but to a lesser extent. As expected, the correspondence accuracy of the landmarks predicted by DCNN-Match Hinge+CE was higher than both DCNN-Match Hinge as well as DCNN-Match CE in all regions of the underlying deformation. Further, the purpose of experimenting with $m_{pos}=0.1$ and $m_{pos}=0.2$ to encourage high landmarks correspondence accuracy in the regions of high deformation seems to be fulfilled. The landmarks correspondence accuracy was high irrespective of the extent of the underlying deformation for DCNN-Match Hinge0.1+CE and even higher for DCNN-Match Hinge0.2+CE.

In Figure \ref{fig:landmarks vs deformation} (b), the spatial density of predicted landmark correspondences (averaged over 121 patients) in different regions of the underlying deformation is plotted for each DCNN-Match variant. The plot shows that DCNN-Match CE predicted more landmarks in regions with high deformations as compared to other DCNN-Match variants, which is purely empirical. 

In Figure \ref{fig:landmarks vs deformation} (c), the $TRE_{after}$ values of the validation landmarks (averaged over 121 patients) in different region of the underlying deformation are plotted for each DCNN-Match variant. As is apparent from the figure, the $TRE_{after}$ values were lowest in all deformation regions when the automatic landmarks predicted by DCNN-Match CE were used as compared to the other DCNN-Match variants.

If we analyze the plots in the Figure \ref{fig:landmarks vs deformation} collectively, we observe that in high deformation regions, DCNN-Match CE predicted landmarks with lower landmarks correspondence accuracy but higher spatial density as compared to DCNN-Match Hinge+CE, DCNN-Match Hinge0.1+CE, and DCNN-Match Hinge0.2+CE. Further, the DIR performance in the highly deformed regions was higher (reflected by lower $TRE_after$ values) with the use of the automatic landmarks predicted by DCNN-Match CE as compared to DCNN-Match Hinge+CE, DCNN-Match Hinge0.1+CE, and DCNN-Match Hinge0.2+CE. This implies that a larger number of slightly less accurate landmarks in highly deformed regions may be more favorable for guiding the DIR approach as compared to a smaller number of highly accurate landmarks.

\begin{figure*}[h!]
    \centering
    \includegraphics[width=\textwidth]{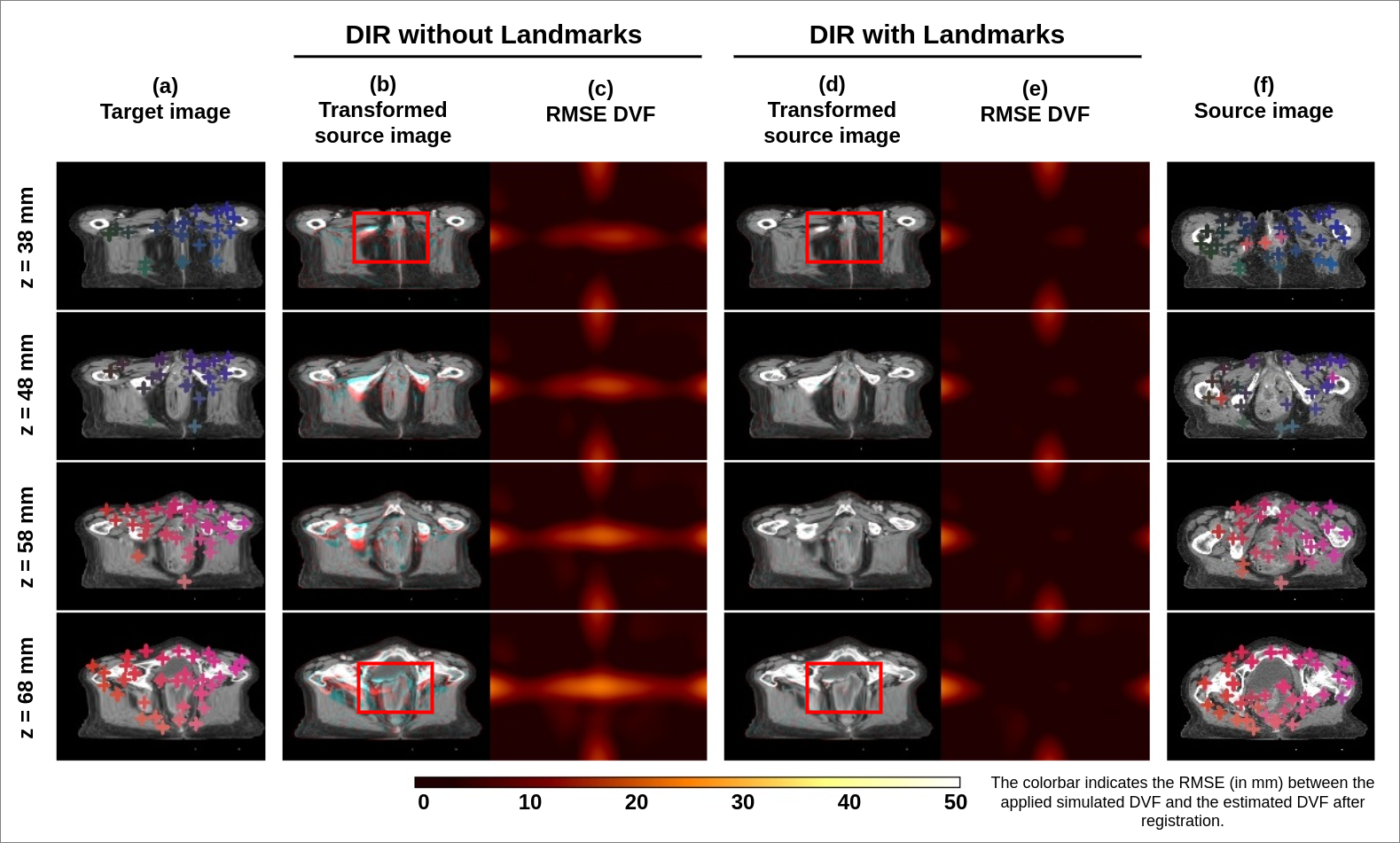}
    \caption{\textbf{Qualitative results on simulated deformations test set.} Transverse slices from 10 mm apart from a representative example are shown in different rows. Column (a): target image, columns (b) and (c): transformed source image and Root Mean Square Error (RMSE) plot between the ground truth and estimated DVF after registration without automatic landmarks, respectively, columns (d) and (e): transformed source image and RMSE plot between the ground truth and estimated DVF after registration with using automatic landmarks predicted by DCNN-Match CE, respectively, column (f): source image. Landmark correspondences between the target and source images are shown in similar colored cross-hairs in columns (a) and (f). Note: some of the landmarks may have correspondences in the transverse slices not shown in the figure. The red rectangles highlight the effect of using landmark correspondences in a highly deformed region.}
    \label{fig:qualitative_simulated}
\end{figure*}

\subsection{Determinant of Spatial Jacobian \& Qualitative Evaluation}

The determinant of the spatial Jacobian of the obtained DVFs was observed to be non-negative in all the registrations obtained in all the experimental scenarios. This indicates that all the obtained registrations were anatomically plausible.
\begin{figure*}[h!]
    \centering
    \includegraphics[width=\textwidth]{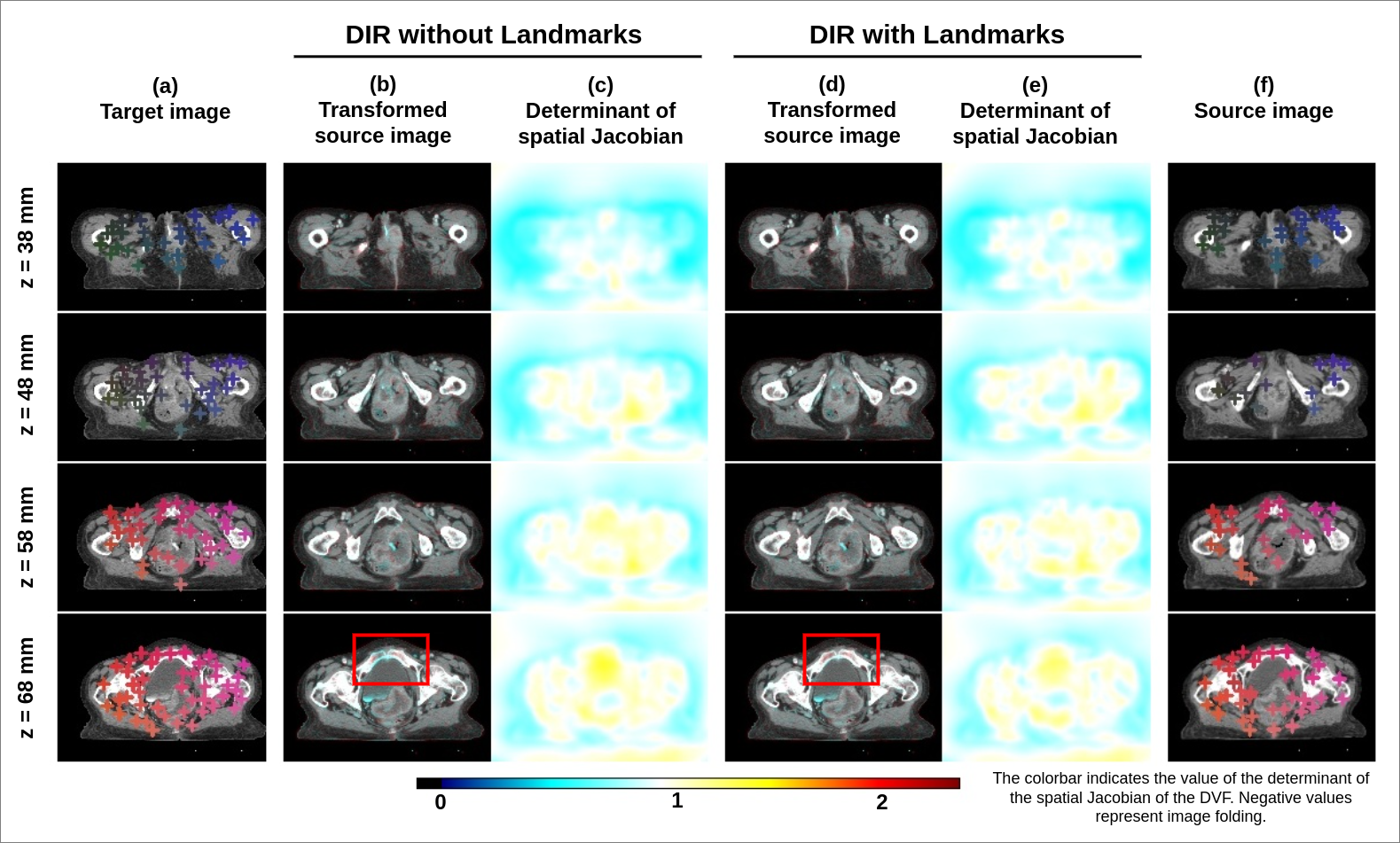}
    \caption{\textbf{Qualitative results on clinical deformations test set.} Transverse slices from 10 mm apart from a representative example are shown in different rows. Column (a): target image, columns (b) and (c): transformed source image and determinant of the spatial Jacobian after registration without automatic landmarks, respectively, columns (d) and (e): transformed source image and determinant of the spatial Jacobian after registration with using automatic landmarks predicted by DCNN-Match CE, respectively, column (f): source image. Landmark correspondences between the target and source images are shown in similar colored cross-hairs in columns (a) and (f). Note: some of the landmarks may have correspondences in the transverse slices not shown in the figure. The red rectangle highlights a region where improvement by adding landmarks correspondences in the DIR is visible.}
    \label{fig:qualitative_clinical}
\end{figure*}

Figure \ref{fig:qualitative_simulated} shows a representative example of registration without using landmarks and registration with the DCNN-Match CE approach. The source image has a large local deformation in the center along with small random deformations globally. The transformed source images obtained after DIR have been overlaid onto the target image (columns (b) and (d)) using complementary colors such that the aligned structures look grey and misalignment is highlighted in colors. As can be seen in column (b), many regions are not aligned properly after the registration, but, with the additional guidance information (column (d)), the anatomical structures look perfectly aligned. The corresponding landmark pairs are shown with cross-hairs of the same color in the target and source images. It is worth noting that DCNN-Match CE can find landmark correspondences in highly deformed regions as well. As a result, DIR with landmark correspondences can find a better estimation of the underlying deformation field as compared to the baseline DIR approach.
Columns (c) and (e) represent the Root Mean Square Errors (RMSE) of the ground truth DVF and the DVF obtained after DIR without and with landmark correspondences.  Further, Figure \ref{fig:qualitative_clinical} shows an example of DIR without and with using landmarks for clinical deformations. While the output of registration without and with using landmark correspondences looks similar in most cases, a subtle improvement in alignment can still be spotted in some regions of the images (also highlighted with a red rectangle in the figure) with the use of landmark correspondences in the DIR. The determinant of the spatial Jacobian shown in Figure \ref{fig:qualitative_clinical} (c) and (e) shows no visible image folding in the DIR solutions obtained by either of the approaches.

\begin{figure}
    \centering
    \includegraphics[width=0.9\textwidth]{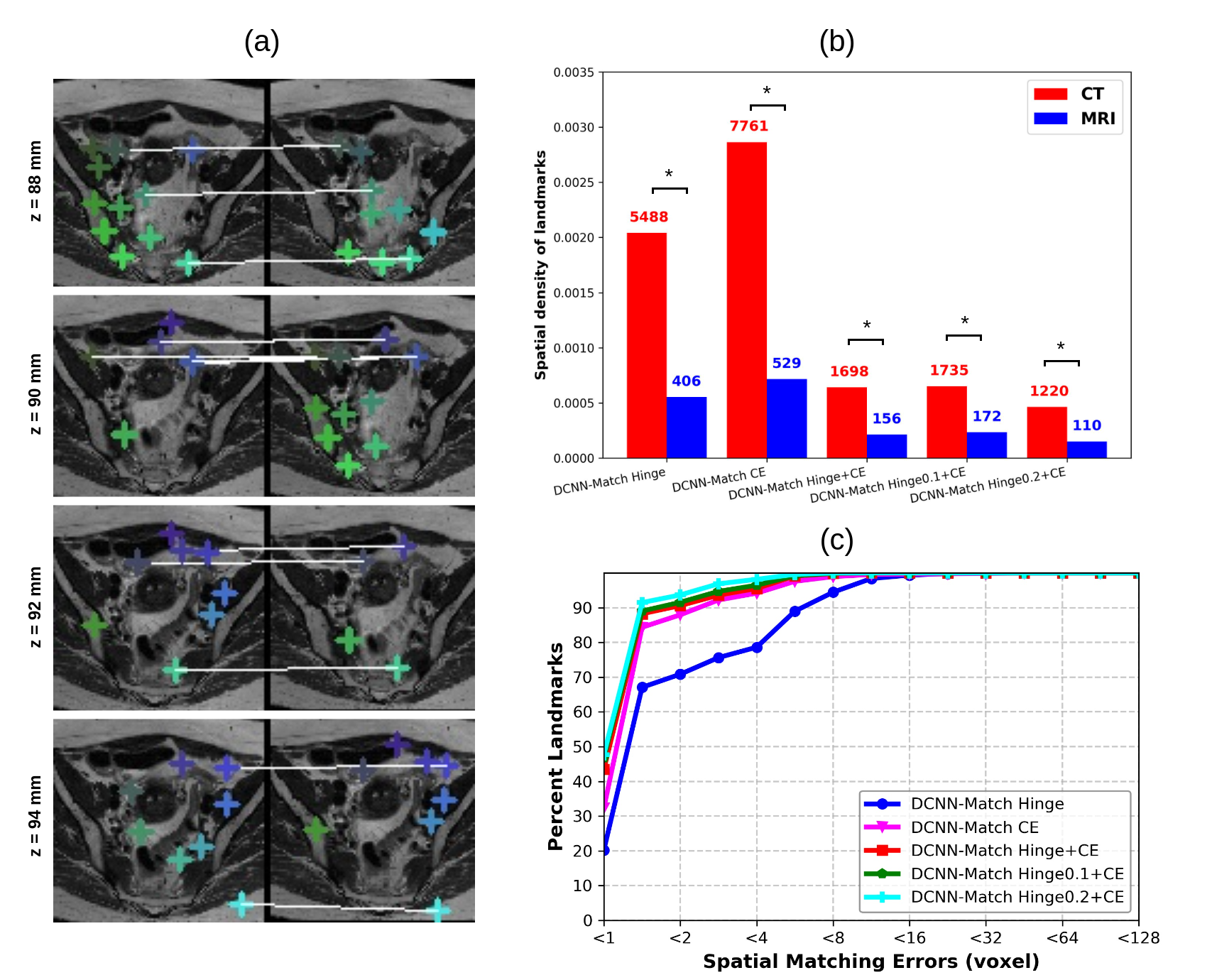}
    \caption{\textbf{Generalization results on the simulated deformations test set - MRI.} (a) Predicted corresponding landmarks in the target and source MRI. Corresponding landmarks are shown with similar colored cross-hairs in the target and source images. Note that some of the landmarks match across slices following the underlying deformation in 3D. (b) Comparison of the spatial density of predicted landmarks (averaged over all patients) between simulated deformations test set - CT and simulated deformations test set - MRI for each DCNN-Match variant. The average number of predicted landmarks is shown in the text above bars. * indicates significant difference after Mann-Whitney U test. (c) Spatial matching errors of predicted landmark correspondences.}
    \label{fig:mri}
\end{figure}

\subsection{Generalization to MRI dataset}
A representative example of predicted landmark correspondences by DCNN-Match CE on MRI scans without retraining is shown in Figure~\ref{fig:mri} (a). Upon visual inspection, the predicted landmark correspondences seem to be accurate despite the different modality of the test scans. Further, the FOV of the acquisition of MRI scans was approximately 16 times smaller than the FOV of the acquisition of CT scans in the test set. To make a direct comparison between the number of predicted landmark correspondences in CT and MRI datasets, we calculated the spatial density of predicted landmarks by dividing the number of landmarks by the total number of voxels in each image. In CT scan images, a large portion of the image consists of background voxels where the DCNN-Match variants do not predict landmark correspondences. Therefore, we considered only the voxels in the patient's anatomy by counting the number of voxels in the largest connected component after binarizing the image through intensity thresholding.

The spatial density of predicted landmarks in both CT and MRI test sets is shown in Figure~\ref{fig:mri} (b). Since the networks were not trained on MRI scans, the spatial density of the predicted landmarks was reduced in MRI scans. Still, a considerable number of landmarks (on average for all patients) were predicted in the MRI test set by each approach (shown as the text above bars). Further, the spatial matching errors (shown in Figure~\ref{fig:mri} (c)) of the predicted landmark correspondences on MRI scans were comparable to the spatial matching errors observed for CT scans. Overall, the above results demonstrate the generalization potential of DCNN-Match on cross-modality data without retraining.

\section{Discussion}
\label{section:discussion}
We developed a self-supervised deep learning method (DCNN-Match) for automatic landmarks correspondence detection in 3D medical images. We have also presented quantitative and qualitative evidence that a high number of landmark correspondences with good spatial matching accuracy can be predicted within seconds with the help of our proposed approach. Furthermore, we integrated DCNN-Match with a DIR pipeline and assessed the added value of automatic landmark correspondences toward the improvement of intra-patient DIR performance. To the best of our knowledge, this is the first study to develop a self-supervised deep learning approach for predicting automatic landmark correspondences in 3D medical images and investigating their applicability in improving DIR.

We developed five variants of the proposed approach, which differed in the way feature descriptor matching is learned. We observed that a separate module for learning feature descriptor matching (DCNN-Match CE) yields landmark correspondences with not only reduced spatial matching errors but also an increased number of matches in regions of high deformation. The results also showed that the added value to the performance of DIR was most prominent by the use of automatic landmark correspondences predicted by DCNN-Match CE. While three other variants predicted automatic landmark correspondences with better spatial matching accuracy than DCNN-Match CE, the numbers of predicted landmarks by these variants were fewer than the number of landmarks predicted by DCNN-Match, especially in regions of high deformation. This implies that the spatial density of predicted landmarks with respect to the underlying deformation plays a role in the extent of the added value provided by the automated landmark correspondences.

The results also showed that the additional guidance by automatic landmark correspondences improved the performance of DIR irrespective of the variance in the number, spatial matching errors, and spatial distribution of the automatic landmarks in both simulated as well as clinical deformations test sets. These findings are in line with the existing literature on the use of automatic landmarks for the improvement of DIR in chest CT \cite{ehrhardt2010automatic, polzin2013combining, ForstnerOperator}, head and neck CT \cite{kearney2014automated}, retinal images \cite{hervella2018multimodal}, and brain MRI images \cite{han2014robust, HAN2015277}. A study on DIR of thoracic CT scans \cite{Alexander2011} reported that automatic landmarks-based optimization of the regularization parameter reduced the TRE of expert landmarks on average by 0.07 mm. Another study on registration of CT scans corresponding to end-inspiration and end-expiration phases reported a reduction of TRE of expert landmarks from  1.34 $\pm$ 2.00 mm to 0.82 $\pm$ 0.97 mm by the use of automatic landmarks in DIR \cite{ForstnerOperator}. Our experiments showed that the TRE of validation landmarks in the simulated deformations test set reduced from 5.07 $\pm$ 9.98 to 3.14 $\pm$ 8.61, and the TRE of expert landmarks in the clinical deformations test set reduced from 6.85 $\pm$ 5.79 to 6.42 $\pm$ 5.79 on average by the use of automatic landmark correspondences predicted by DCNN-Match CE in DIR. Since the improvement in DIR performance reported in terms of TRE values of the expert landmarks is affected by several factors e.g., the number and location of the expert landmarks, image resolution, and TRE values before registration, a comparison in absolute values of TRE improvement cannot be made.
Nevertheless, the current study adds to the existing evidence on the added value of automatic landmark correspondences in improving DIR by providing experimental results from pelvic CT scan registrations, which otherwise did not exist.

Two other studies have looked into intra-patient DIR in cervical cancer patients \cite{rigaud2019deformable, bondar2010symmetric}. The authors in one of the studies \cite{rigaud2019deformable} have focused on dose mapping and do not report TRE values. The average TRE values after registration reported in the other study \cite{bondar2010symmetric} are the following: 3.5 $\pm$ 2.4 mm for bladder top, 8.5 $\pm$ 5.2 mm for cervix tip, 5.7 $\pm$ 2.1 mm for markers, and 4.6 $\pm$ 2.2 mm for the midline. As such, a direct correspondence between the landmarks used in our study and landmarks in the earlier study cannot be ascertained. Moreover, the underlying dataset and methods used are also different. Still, the mean TRE value obtained after registration with additional guidance information from landmark correspondences predicted by DCNN-MatchCE (6.42 $\pm$ 5.79 mm) seems to be within the range of reported TRE values, which gives some confidence that the obtained DIR results are satisfactory.

The extent of the added value provided by the use of automatic landmark correspondences in DIR was lower in the clinical deformations test set as compared to the simulated deformations test set. Our retrospective analysis (provided in supplementary material S1) revealed no obvious patterns regarding the spatial distribution of the automatic landmarks in relation to manual landmarks used for TRE calculations that could explain the lower added value of using automatic landmarks in the clinical deformations test set. The DIR performance in case of clinical deformations as reflected by TRE of manually annotated landmarks is affected by several factors e.g., choice of manual landmarks, inter- and intra-observer variation in the placement of manual landmarks, hyperparameters in the parameter map used for Elastix, limitations of Elastix in modeling large deformations, sliding tissue, and singularities in DVF. Therefore, establishing a direct relationship between the quality of automatic landmark correspondences and the DIR performance is difficult. However, we can speculate on a few factors that impacted the quality of automatic landmark correspondences and hence could have impacted the added value to DIR. In the clinical test set, the CT scans were acquired with contrast administered via a rectal tube or intravenously. Consequently, one or multiple regions (e.g., vagina, bladder, bowel bag, or vascular regions) were contrast-enhanced giving rise to large differences in appearance between the CT scan pairs, which was not a part of the training for DCNN-Match. An example of appearance variation due to contrast is shown in Figure \ref{fig:contrast example} (b). This appearance variance between the source and target CT scans often overlapped with the large and complex deformations in the bladder and bowel bag. This posed an additional challenge for finding landmark correspondences between scans. Although all DCNN-Match variants were still able to find landmark correspondences in these scans despite the aforementioned challenges, they failed to find correspondences in regions where appearance was strongly different due to a combination of contrast administration and underlying deformation. We expect that incorporating a model for simulating contrast differences between scans and a better (probably a bio-mechanical based) model for simulating deformations due to physical phenomena such as bladder filling would lead to the prediction of automatic landmarks in the aforementioned challenging scenario as well and yield a larger added value of using automatic landmark correspondences in DIR. We are considering pursuing this direction for a future study.

Another factor affecting the DIR performance in the clinical deformations test set is that we tuned the hyperparameters used in Elastix (weights of the objectives used in DIR, $weight_1$, and $weight_2$) based on the DIR of CT scan pairs in the validation set consisting of simulated deformations. We used these hyperparameters for all the registrations in both simulated as well as clinical deformation test sets. This does not acknowledge the fact that each DIR problem is unique and therefore, a single setting for all source and target pairs is sub-optimal. Earlier research has also pointed out the importance of tuning the weights of different objectives in the DIR separately for each image pair to achieve the best DIR performance \cite{Alexander2011, pirpinia2017feasibility}. We conducted retrospective experiments by changing the weights of the objectives in DIR, which revealed that $weight_1$, and $weight_2$ values corresponding to best DIR performance (quantified in terms of minimum TRE values) were indeed different for each CT scan pair in the clinical deformations test set. Unfortunately, the tuning of $weight_1$, and $weight_2$ separately for each CT scan pair in the clinical deformations test set could not be done objectively and automatically due to the unavailability of the underlying ground truth. Note that the manually annotated landmarks were used to evaluate the DIR performance and therefore using them for tuning $weight_1$, and $weight_2$ would have produced biased results. However, the purpose of this research was not to obtain the best DIR performance for each CT scan pair but to quantify the effect of additional guidance provided by the automatic landmark correspondences. 

Further, the added value of the additional guidance provided by the automatic landmark correspondences may be limited by erroneous matches. While the results on the simulated data indicated the benefits of more landmarks toward DIR performance, the adverse effect of erroneous matches remains unclear. It would be interesting to investigate in a future study how much value can be gained by removing the erroneous landmark matches either using RANSAC \cite{fischler1981random} or a deep learning approach \cite{yi2018learning}. Another interesting direction for future research can be to simultaneously learn a deep learning model for landmark matching as well as performing DIR. Such a model can be used to investigate how many landmarks are optimal for improving the DIR. However, care needs to be taken to avoid degeneracy because landmark matching essentially is performing DIR on a sparse grid and the optimal number of landmark matches to improve DIR could quite likely be the total number of voxels in the image. 

Remarkably the proposed approach for finding automatic landmark correspondences could find automatic landmark correspondences on cross-modality data without retraining. Based on this observation, we expect that with retraining (which requires minimal effort because manual annotations are not needed), the proposed approach should be able to find automatic landmark correspondences on any type of medical imaging data. Furthermore, since a considerable number of landmarks were predicted in the MRI scans with spatial matching errors comparable to the CT scans, we expect that the use of automatic landmarks should lead to performance gain in DIR on MRI scans also. With retraining on MRI scans, we expect that the added value to the DIR performance will be similar to as observed in the CT scans.

\section{Conclusion}
\label{section:conclusion}

We developed a self-supervised method for automatic landmarks correspondence detection in abdominal CT scans and investigated the effect of different variants of our automatic landmarks correspondence detection approach on the performance of DIR. The obtained results provide strong evidence for the added value of using automatic landmark correspondences in providing additional guidance information to DIR. The added value of automatic landmarks in DIR is consistent across different variants of our approach and for both simulated as well as clinical deformations. Additionally, we observed that the spatial distribution of automatic landmark correspondences with respect to the underlying deformation has a considerable effect on the extent of the added value provided by landmark correspondences. A higher number of automatic landmark correspondences in highly deformed regions has more added value than more accurate but fewer landmark correspondences. Therefore, further research in the direction of developing landmark detection approaches that are aware of the underlying deformation is recommended. 

In conclusion, the current study affirms the added value of using automatic landmark correspondences for solving challenging DIR problems and provides insights into what type of landmark correspondences (in terms of spatial distribution and matching errors) may be more beneficial to DIR than others.

\newpage
\appendix    
\label{sec:supplementary material}
\section{Elastix Parameter Maps}
\label{appendix: elastix parameter maps}
\subsection{Affine Registration}
\begin{lstlisting}
(AutomaticParameterEstimation "true")
(AutomaticTransformInitialization "true")
(AutomaticTransformInitializationMethod "Origins")
(CheckNumberOfSamples "true")
(DefaultPixelValue 0)
(FinalBSplineInterpolationOrder 1)
(FixedImagePyramid "FixedSmoothingImagePyramid")
(ImageSampler "RandomCoordinate")
(Interpolator "LinearInterpolator")
(MaximumNumberOfIterations 1024)
(MaximumNumberOfSamplingAttempts 8)
(Metric "AdvancedMattesMutualInformation")
(MovingImagePyramid "MovingSmoothingImagePyramid")
(NewSamplesEveryIteration "true")
(NumberOfResolutions 4)
(NumberOfSamplesForExactGradient 4096)
(NumberOfSpatialSamples 4096)
(Optimizer "AdaptiveStochasticGradientDescent")
(Registration "MultiResolutionRegistration")
(ResampleInterpolator "FinalBSplineInterpolator")
(Resampler "DefaultResampler")
(Transform "AffineTransform")
\end{lstlisting}

\subsection{Deformable Image Registration}
\begin{lstlisting}
(AutomaticParameterEstimation "true")
(BSplineInterpolationOrder 1)
(CheckNumberOfSamples "true")
(DefaultPixelValue 0)
(FinalBSplineInterpolationOrder 1)
(FinalGridSpacingInPhysicalUnits 8)
(FixedImageDimension 3)
(FixedImagePixelType "float")
(FixedImagePyramid "FixedRecursiveImagePyramid")
(HowToCombineTransforms "Compose")
(ImageSampler "RandomCoordinate")
(Interpolator "BSplineInterpolator")
(MaximumNumberOfIterations 300 600 900 1200)
(Metric "AdvancedMattesMutualInformation" "TransformBendingEnergyPenalty"
        "CorrespondingPointsEuclideanDistanceMetric")
(Metric0Weight 1)
(Metric1Weight 1)
(Metric2Weight 0.01)
(MovingImageDimension 3)
(MovingImagePixelType "float")
(MovingImagePyramid "MovingRecursiveImagePyramid")
(NewSamplesEveryIteration "true" "true" "true" "true")
(NumberOfHistogramBins 32 32 32 32)
(NumberOfResolutions 4)
(NumberOfSpatialSamples 5000 5000 5000 5000)
(Optimizer "StandardGradientDescent")
(Registration "MultiMetricMultiResolutionRegistration")
(ResampleInterpolator "FinalBSplineInterpolator")
(Resampler "DefaultResampler")
(SP_A 100 200 300 400)
(SP_a 35000 30000 25000 20000)
(SP_alpha 0.602 0.602 0.602 0.602)
(ShowExactMetricValue "false" "false" "false" "false")
(Transform "BSplineTransform")
(UpsampleGridOption "true")
\end{lstlisting}

\section{List of manually annotated landmarks}
\label{appendix: landmarks list}
\begin{itemize}
    \item fiducial markers in the vaginal wall near the cervix at the locations: posterior left, anterior mid, posterior right, posterior mid, anterior left, and anterior right
    \item bifurcation aorta
    \item os coccygis
    \item medial tip of right and left trochanter minor
    \item most caudal, dorsal, and ventral part of the corpus of lumbar vertebrae 3
    \item most caudal, dorsal, and ventral part of the corpus of lumbar vertebrae 5
    \item right and left bifurcation vena iliaca communis
    \item right and left bifurcation of artery iliaca communis
    \item umbilicus
    \item caudal tip of right and left kidney
    \item external and internal anal sphincter
    \item cervical ostium
    \item external and internal urethral ostium
    \item right and left ureteral ostium
    \item uterus top
\end{itemize}

\subsection* {Code, Data, and Materials Availability} 
The code for DCNN-Match is available at: \href{https://github.com/monikagrewal/End2EndLandmarks}{End2EndLandmarks\_repo}.

\subsection*{Disclosures}
This work was presented (in part) at the Conference of SPIE Medical Imaging 1131303: Image-Guided Procedures, Robotic Interventions, and Modeling (February 15 to February 20, 2020, Houston, Texas, USA).

This work was supported by Elekta (Elekta AB, Stockholm, Sweden) and Xomnia (Xomnia B.V., Amsterdam, the Netherlands). Elekta and Xomnia were not involved in the study design, data collection, analysis and interpretation, and writing of this article.

The authors have no relevant financial interests in the manuscript.
The authors have no conflict of interests.

\subsection* {Acknowledgments}
The research is part of the research programme, Open Technology Programme with project number 15586, which is financed by the Dutch Research Council (NWO), Elekta, and Xomnia. Further, the work is co-funded by the public-private partnership allowance for top consortia for knowledge and innovation (TKIs) from the Ministry of Economic Affairs. 


\bibliography{report}   
\bibliographystyle{spiejour}   


\vspace{2ex}\noindent\textbf{Monika Grewal} is a PhD student in the Evolutionary Intelligence research group at Centrum Wiskunde \& Informatica, Amsterdam, The Netherlands. She received her B.Tech and M.Tech degrees in Electronics \& Communication Engineering in 2010 and 2012, respectively. The focus of her PhD research is to develop DIR methods for advancing and simplifying the radiotherapy treatment workflow, especially for cervical cancer patients. Her research interests include artificial intelligence (specifically deep learning and evolutionary algorithms) and medical imaging.

\vspace{2ex}\noindent\textbf{Peter A. N. Bosman} is the group leader of the Evolutionary Intelligence research group at the Dutch National Research Institute for Mathematics and Computer Science (Centrum Wiskunde \& Informatica) and a professor of evolutionary algorithms at Delft University of Technology. His research concerns the design of scalable model-based evolutionary algorithms and their application, primarily in the life sciences and health domain. He has (co-)authored more than 200 refereed publications, out of which 8 received best paper awards.

\vspace{2ex}\noindent\textbf{Tanja Alderliesten} is an associate professor in the Department of Radiation Oncology, Leiden University Medical Center, the Netherlands, where she is the group leader of the Artificial Intelligence-based Innovations research group. The focus of her research is translational in nature and primarily concerns the development of state-of-the-art methods and techniques from the fields of mathematics and computer science (including image processing, biomechanical modeling,  optimization, and (explainable) artificial intelligence) for (radiation) oncology.

\newpage
\vspace{2ex}\noindent\textbf{Dr. Henrike Westerveld} is a radiation oncologist at the Erasmus Medical Center, Rotterdam, the Netherlands. She obtained her MD from the University of Amsterdam in 2001, and her PhD in 2008. She has worked at the AKH in Vienna (A), the UMCU in Utrecht (NL) and the AMC (now AmsterdamUMC)(NL). Dr. Westerveld is specialized in the treatment of gynecological and urological tumors. Her main interest is in image-guided radiotherapy and late effects in gynecological cancers.

\vspace{1ex}
\noindent Biographies and photographs of the other authors are not available.

\listoffigures
\listoftables

\end{spacing}
\end{document}


\maketitle

\noindent \journal{Journal of Medical Imaging}
\manuscript{Automatic Landmarks Correspondence Detection in Medical Images with an Application to Deformable Image Registration}

\section*{Retrospective Analysis}
The extent of the added value provided by the use of automatic landmark correspondences in DIR was lower in the clinical deformations test set as compared to the simulated deformations test set. Therefore, we analyzed the TRE values of each manual landmarks in the clinical deformations test set to understand the possible causes for the lack of performance gain by using automatic landmarks in DIR.
\begin{figure*}[h!]
    \centering
    \includegraphics[width=\textwidth]{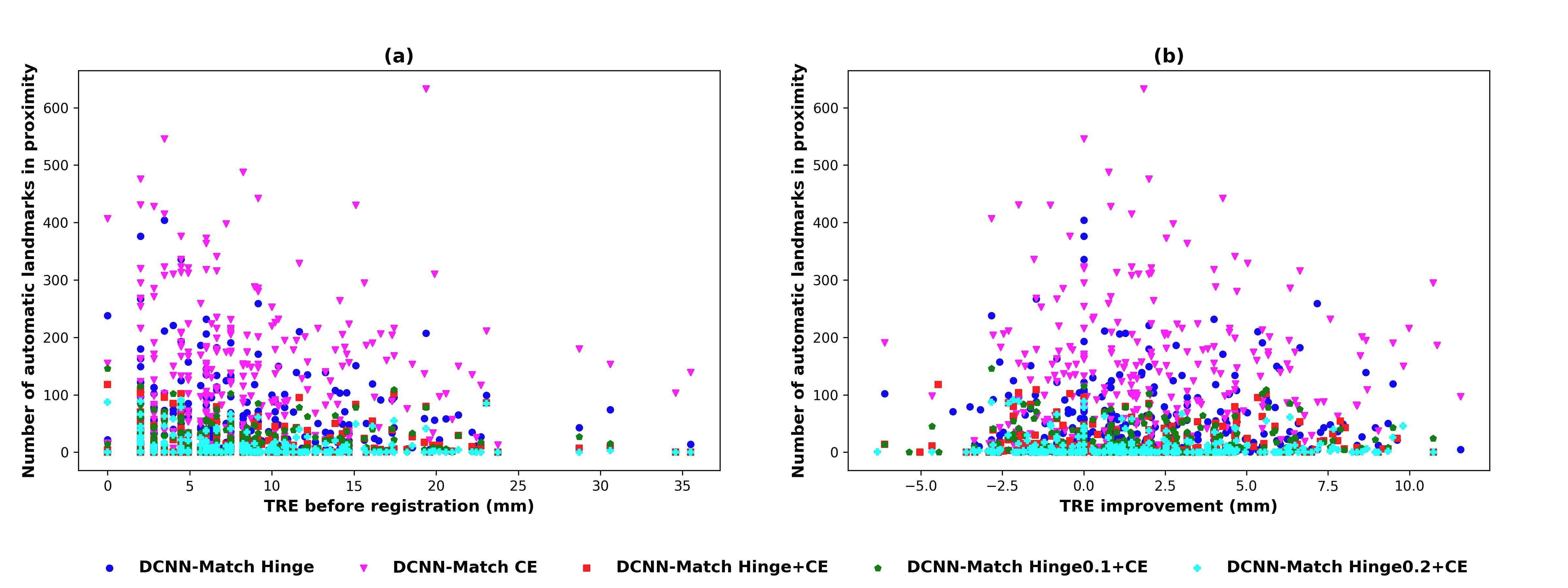}
    \caption{\textbf{Results on clinical deformations test set.} For each manual landmark, the number of automatic landmark correspondences predicted in 16 mm proximity to that manual landmark has been plotted against (a) the corresponding $TRE_{before}$ value and (b) TRE improvement value as obtained by subtracting $TRE_{after}$ value from $TRE_{before}$ value.}
    \label{fig:landmark_vs_trebefore}
\end{figure*}
Specifically, we calculated the number of automatic landmarks in proximity (16 mm) to each of the manual landmark. We plotted this value against the $TRE_{before}$ value (representative of the underlying deformation in that region) of that manual landmark (shown in Figure \ref{fig:landmark_vs_trebefore} (a)). The plot shows that automatic landmarks were predicted in the regions of high deformation as well, especially by DCNN-Match CE. Therefore, a lack of the presence of automatic landmarks in highly deformed regions could not be the sole cause for the lack of performance gain in DIR.

Further, we calculated the TRE improvement for each manual landmark by subtracting $TRE_{after}$ from $TRE_{before}$ values. A positive high number indicates higher improvement in TRE values (or DIR performance). In Figure \ref{fig:landmark_vs_trebefore} (b), the TRE improvement values have been plotted against the number of automatic landmarks in proximity for each manual landmark. We observed that the TRE value of some of the manual landmarks in some of the patients did not improve despite the presence of automatic landmarks in their proximity.

In conclusion, the above analysis shows that a straightforward pattern regarding the spatial distribution of automatic landmarks relative to the manual landmarks cannot be established in case of clinical deformations. Consequently, a direct relationship between the quality of automatic landmark correspondences and the DIR performance cannot be established.